\documentclass{article} 
\usepackage{iclr2025_conference,times}

\usepackage{amsmath,amsfonts,bm}

\def\eqref#1{equation~\ref{#1}}

\def\1{\bm{1}}

\DeclareMathAlphabet{\mathsfit}{\encodingdefault}{\sfdefault}{m}{sl}
\SetMathAlphabet{\mathsfit}{bold}{\encodingdefault}{\sfdefault}{bx}{n}

\newcommand{\sigmoid}{\sigma}

\usepackage{hyperref}
\usepackage{cleveref}
\usepackage{comment}        
\usepackage{subcaption}
\usepackage{booktabs}
\usepackage{multirow}
\usepackage{dirtytalk}
\usepackage{url}
\usepackage{graphicx}
\usepackage{float}
\usepackage{tikz}
\newcommand*\circled[1]{\tikz[baseline=(char.base)]{
            \node[shape=circle,draw,inner sep=0.7pt] (char) {#1};}}

\title{Uncertainty-Penalized Direct Preference \\Optimization}

\usepackage{authblk}

\author[1]{Sam Houliston\thanks{Correspondence to: \href{mailto:shouliston@student.ethz.ch}{\texttt{shouliston@student.ethz.ch}}.}$^{*,}$}
\author[1,2,3]{Alizée Pace}
\author[1,3]{Alexander Immer}
\author[1,2]{Gunnar Rätsch}
\affil[1]{Department of Computer Science, ETH Zurich, Switzerland}
\affil[2]{ETH AI Center, Zurich, Switzerland}
\affil[3]{Max Planck Institute for Intelligent Systems, T\"ubingen, Germany}

\iclrfinalcopy 
\begin{document}

\maketitle

\begin{abstract}
Aligning Large Language Models (LLMs) to human preferences in content, style, and presentation is challenging, in part because preferences are varied, context-dependent, and sometimes inherently ambiguous. While successful, Reinforcement Learning from Human Feedback (RLHF) and Direct Preference Optimization (DPO) are prone to the issue of proxy reward overoptimization. 
Analysis of the DPO loss reveals a critical need for regularization for mislabeled or ambiguous preference pairs to avoid reward hacking.  In this work, we develop a pessimistic framework for DPO by introducing preference uncertainty penalization schemes, inspired by offline reinforcement learning. 
The penalization serves as a correction to the loss which attenuates the loss gradient for uncertain samples.
Evaluation of the methods is performed with GPT2 Medium on the Anthropic-HH dataset using a model ensemble to obtain uncertainty estimates, and shows improved overall performance compared to vanilla DPO, as well as better completions on prompts from high-uncertainty chosen/rejected responses.
\end{abstract}

\section{Introduction}

Aligning LLMs to human preferences in content, style, and presentation has become a central challenge in improving and deploying LLMs, leading to the advent of Reinforcement Learning with Human Feedback (RLHF), now a prominent technique to fine-tune state-of-the-art LLMs \citep{openproblemsinRLHF}. The standard RLHF pipeline involves human feedback collection, reward model training, and LLM policy optimization via reinforcement learning (RL). Despite its success, each stage presents challenges, from feedback interpretation and policy generalization to challenging RL implementation \citep{openproblemsinRLHF}. 
Direct Preference Optimisation (DPO) \citep{OGDPO} effectively bypasses the reward model by fine-tuning the policy to maximize the likelihood of the preference data under the Bradley–Terry model \citep{bradleyT}. DPO is easier to implement than RL algorithms, and benefits from computational efficiency and stability by avoiding potential inaccuracies and biases of a reward model \citep{dposuperiorppo, openproblemsinRLHF}.

A predominant issue in RLHF techniques is proxy reward overoptimization, arising from the assumption that preferences are effectively captured as pointwise rewards (or binary comparisons), or from an imperfect coverage of the full preference distribution by the training data. In particular, DPO easily overfits on the preference dataset \citep{GeneralParadigmforPreferences}, and considers all binary preference pairs in the dataset equally, while some may be stronger than others \citep{offsetdpo}. We study the DPO loss to understand this overfitting regime in \Cref{section:dpolossanalysis} and reveal that DPO is particularly sensitive to erroneous or ambiguous preference pairs.

Inspiration can be taken from Offline RL which learns a policy from a fixed dataset but suffers from the distributional shift problem, where reward or value overestimation errors occur when the policy encounters OOD states that are underrepresented in the dataset, leading to poor generalization. To address this, pessimism towards OOD states is induced by penalizing high uncertainty rewards, which encourages reliable policy learning \citep{pessimisticRL, pessimisticRLdealingwiththeunknwown}. 
RLHF shares similar issues with offline RL, and may benefit from additional reward uncertainty quantification as information on the validity of preference pairs.


In this work, we introduce uncertainty penalization schemes for DPO inspired by pessimistic offline RL. 
Our best-performing scheme multiplies implicit rewards by an energy-based function of the uncertainty -this scheme is tailored to the DPO loss and derived in \Cref{section:method}. Our framework assumes access to uncertainty estimates on the preferences, which could be obtained upon data collection or through an additional reward model equipped with uncertainty quantification.

Our main contributions are as follows. (1) We provide a fine-grained analysis of DPO and its overfitting problem. Our analysis complements the existing literature \citep{GeneralParadigmforPreferences, DPOP, offsetdpo, filtereddpo} and highlights the sensitivity of DPO to mislabeled samples. (2) We introduce a pessimistic framework that addresses DPO's overfitting problem through uncertainty penalization. Our framework is general to RLHF, but specialized to DPO variants such as IPO \citep{GeneralParadigmforPreferences}. The penalization schemes are inspired by the established techniques of pessimism in offline reinforcement learning. (3) We demonstrate the superior performance of our methods over unpenalized objectives on a range of tasks and robustness experiments.

\section{Related Work}\label{section:relatedwork}

\subsection{Uncertainty Penalization in Standard RLHF}
\citet{LORARLHF} show empirically that Kullback-Leibler (KL) regularization in the standard RLHF pipeline may be insufficient to avoid reward overoptimization. Their method Uncertainty Penalized-RLHF (UP-RLHF) trains an ensemble of diverse Low-Rank-Adaptation (LoRA) reward models to obtain a mean reward and ensemble epistemic uncertainty. A Lower Confidence Bound (LCB) on the reward is taken by subtracting a factor of the uncertainty from the mean. Similarly, \citet{bayesianreward} employ a bayesian reward model to use conservative reward estimates in the best-of-n sampling framework.

\subsection{DPO and Variants}
DPO \citep{OGDPO} is an effective approach to finetuning for binary preferences without a reward model or reinforcement learning, while still optimizing for the original RLHF objective. The works below study its limitations and extend the method to more involved frameworks.


\cite{GeneralParadigmforPreferences} notice DPO easily overfits on training preferences, especially for inputs where the policy's implicit reward are nearly deterministic (close to 1 or 0). They introduce Identity Preference Optimisation (IPO) which adds a regularisation term to DPO, enabling one to train models to convergence without requiring tricks like early stopping. In addition, they unify RLHF, DPO, and IPO under a common mathematical formulation $\Psi$PO.

\cite{DPOP} frame LLM generation as a Markov Decision Process (MDP) at the token level (instead of a contextual bandit at the entire completion level) to show a failure mode of DPO on low-edit-distance preference pairs. In this case, DPO increases the relative probability between the chosen and rejected text, however is a reduction in both absolute likelihoods. DPO-Positive (DPO-P) adds a clipping term to the loss to ensure positive log-likelihood of the chosen text. 

\cite{offsetdpo} introduce Offstet DPO (ODPO) which adds a margin between the implicit chosen/rejected rewards in the DPO loss. The margin is based on reward scores given by an external reward model, to help DPO distinguish between strong or weak preference pairs. 
Our work is similar to ODPO. For the standard Lower Confidence Bound uncertainty penalization, we recover a similar loss formulation with a margin (\ref{dpoobjectivepenalized}); our margin equals the difference in chosen-rejected uncertainties, whereas ODPO uses the difference in reward scores. For our main method \textit{Energy Factor Penalization}, the resulting penalized loss does not have an additive margin, instead, the individual chosen-rejected implicit rewards are multiplied by an energy function of their uncertainty (\ref{eq:multiplicationpenalization}) which prevents uncertainties from canceling out, leading to a more precise penalization and improved results. 

Active-DPO by \citet{activedpo} deploys an active-learning strategy to select data for DPO. The acquisition function chooses preference pairs exhibiting the highest predictive entropy of LLM policy uncertainty, to accent the learning on uncertain samples policy-wise.

\citet{chisquared} address overoptimization in DPO by replacing the KL-divergence in the original RLHF objective with the $\chi^2$-divergence, yielding a similar but more robust DPO objective. A sample complexity analysis shows $\chi^2$ regularization offers improved sample efficiency for both offline, online, and mixed setups.

The related works close to our method that involve uncertainty estimation or the inclusion of pessimism into RLHF are summarized in \cref{tab:my-table}. More related works are described in \Cref{app:extendeddporelatedworks}.

\begin{table}[h]
\centering
\renewcommand{\arraystretch}{1.3}
\resizebox{\textwidth}{!}{%
\begin{tabular}{@{}lp{3.5cm}p{4.8cm}c@{}}
\toprule
\multicolumn{1}{c}{\textbf{Method}}       & \multicolumn{1}{c}{\textbf{Uncertainty Quantification Method}}                             & \multicolumn{1}{c}{\textbf{Penalization Scheme}}                                                      & \multicolumn{1}{c}{\textbf{Alignment Algorithm}}  
\\ \midrule
UP-RLHF \citep{diverselora}  
& Diverse LoRA Ensemble                                                               & $r(x,y) \leftarrow r(x,y) - u(y|x)$                                                 & PPO                                               
\\
\citep{bayesianrwdmodel}                                     
& Bayesian LoRA Model 
& $r(x,y) \leftarrow r(x,y) - u(y|x)$               
& Best of N     
\\
Active-DPO \citep{activedpo}
& LLM policy entropy and \newline Implicit rewards margin & Select preference pair from \newline dataset if high $\pi_\theta$ uncertainty & DPO               
\\ \hline
\textbf{Uncertainty Energy Factor (Ours)} 
& LoRA Ensemble
& $\hat{r}_\theta(x,y) \leftarrow \hat{r}_\theta(x,y)\, e^{-u(y|x)/\tau}$
& DPO, IPO
\\ \bottomrule
\end{tabular}%
}\vspace{-0cm}
\caption{Related methods in RLHF that leverage uncertainty estimation.}
\label{tab:my-table}
\end{table}

\section{Analysis of the DPO Loss}\label{section:dpolossanalysis}
Analysis of the DPO loss shows larger policy gradient update steps are applied on preference pairs with low chosen/rejected likelihood ratios compared to reference ratios \ref{eq:advantage}. This behavior can be harmful for mislabeled or similar pairs. Additionally, the overoptimization phenomenon is shown to happen for rejected samples with low probability ($\pi_\theta(y_l|x) \ll 1$), where DPO does not regularize and decreases $\pi_\theta(y_l|x)$ further, potentially increasing the relative probability of other completions. These observations, tied to the overoptimization problem addressed by IPO \citep{GeneralParadigmforPreferences}, motivate the incorporation of pessimism as a gradient update attenuation mechanism for erroneous, or similar (uncertain) samples.

\paragraph{Problem Setting.}
The aim is to align the parametrized LLM policy $\pi_\theta$, to the preference dataset $\mathcal{D} = \{x_i, y_{i,w}, y_{i,l}\}_{i=1}^{N}$ composed of prompts $x$, chosen completions $y_w$, and rejected completions $y_l$, while keeping close to a given reference model policy $\pi_\text{ref}$. The DPO loss \citep{OGDPO} is formulated as:
\begin{align}\label{eq:dpoobjective}
\mathcal{L}_{\text{DPO}}(\pi_\theta; \pi_{\text{ref}})
&=
\mathop{\mathbb{E}}_{(x,y_w,y_l)\sim D} \left[- \log \sigma \left( 
\beta \log \frac{\pi_\theta(y_w | x)}{\pi_{\text{ref}}(y_w | x)} - \beta \log \frac{\pi_\theta(y_l | x)}{\pi_{\text{ref}}(y_l | x)} 
\right) \right]
\\
&=
\mathop{\mathbb{E}}_{(x,y_w,y_l)\sim D} 
\biggl[ - \log \sigma \biggl(
\beta \log 
\underbrace{ \left (
\frac{\pi_\theta(y_w | x) \pi_{\text{ref}}(y_l | x)}{\pi_\theta(y_l | x)\,\pi_{\text{ref}}(y_w | x)}
\right )}_{A_\theta}
\biggr) \biggr].
\end{align}
\vspace{-0.6cm}
\begin{figure}[H]
    \centering
    \begin{minipage}[]{0.72\textwidth}
By the monotonicity of the log and sigmoid functions, minimizing the DPO loss in \Cref{eq:dpoobjective} corresponds to maximizing $A_\theta$. $A_\theta$ is proportional to the policy's chosen-rejected likelihood ratio and provides a measure of how well the policy distinguishes between the completions compared to the reference.
    Figure \ref{fig:dpolossdynamic} shows the loss value and its gradient both increase as $A_\theta$ decreases; thus, the DPO loss severely penalizes (and strongly updates on) inputs where $A_\theta$ approaches zero.
    \end{minipage}%
    \hfill
    \begin{minipage}[]{0.22\textwidth}
        \centering
        \includegraphics[width=\textwidth]{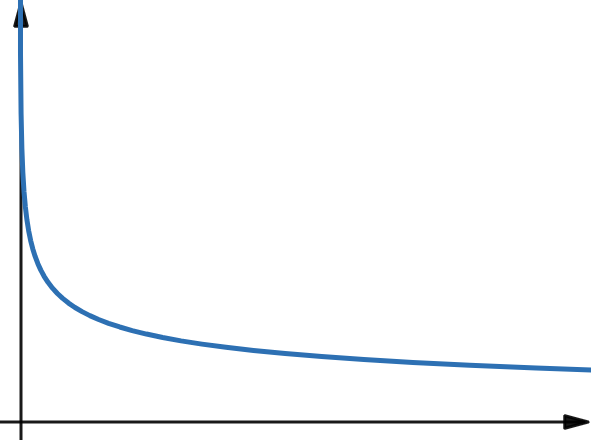}
        \vspace{-0.65cm}
        \caption{$\mathcal{L}_{\text{DPO}}$ vs $A_\theta$}
        \label{fig:dpolossdynamic}
    \end{minipage}
\end{figure}
\vspace{-0.4cm}
 The strong update regime is described in \Cref{eq:advantage} and confirms DPO performs stronger updates when the target policy performs worse relative to the reference policy.
\begin{align}\label{eq:advantage}
    A_\theta = \frac{\pi_\theta(y_w | x)}{\pi_\theta(y_l | x)} \frac{\pi_{\text{ref}}(y_l | x)}{\pi_{\text{ref}}(y_w | x)} \ll 1 
    \Longleftrightarrow \;
    \frac{\pi_\theta(y_w | x)}{\pi_\theta(y_l | x)} 
    \ll
    \frac{\pi_{\text{ref}}(y_w | x)}{\pi_{\text{ref}}(y_l | x)}.
\end{align}
\paragraph{Impact of Ill-Labeled Preference Pairs.} Suppose 
$\tilde{y}_w, \tilde{y}_l$ represent a corrupted or ill-labeled preference pair i.e. $p^*(\tilde{y}_w  \succ \tilde{y}_l) \leq 0.5$ as per the Bradley-Terry model $p^*$. Assuming a decent reference model, we can expected the DPO-trained policy $\pi_\theta$ to perform worse than the reference, and to satisfy the rightmost inequality of \Cref{eq:advantage}. Furthermore, suppose the target $\pi_\theta$ is overfit during training, and that preference pair $\tilde{y}_w, \tilde{y}_l$ is inherently ambiguous $(p^*(\tilde{y}_w  \succ \tilde{y}_l) \approx 0.5)$, it is also likely for the target policy to perform worse than the untrained reference (\Cref{eq:advantage}), and find itself in the strong-update regime. Thus DPO is prone to overfitting on ill-labeled preference pairs, and once overfit, is prone to performing strong updates for inherently ambiguous preference pairs.


\subsection{Analysis of the Gradient}
The DPO loss gradient w.r.t. policy parameters $\theta$ is derived to have a finer understanding of the mechanics of DPO updates that follow gradient-based optimization. The full derivation is provided in \Cref{app:dpoderivation}. For shorthand, we denote the variable $\rho_\theta := \hat{r}_\theta(y_w|x) - \hat{r}_\theta(y_l|x) = \beta \log A_\theta$, as the difference between implicit rewards, which increases with $A_\theta$.
\begin{align}\label{eq:dpograd}
\nabla_\theta \, \mathcal{L}_{\text{DPO}}(x, y_w, y_l; \theta)
&=
\mathop{\mathbb{E}}_{(x,y_w,y_l)\sim D} 
\biggl[
- \beta \underbrace{\sigma(- \rho_\theta)}_{\text{\circled{1}}} 
\underbrace{\left( 
\frac{\nabla_\theta \pi_\theta(y_w | x)}{\pi_\theta(y_w | x)} - \frac{\nabla_\theta \pi_\theta(y_l | x)}{\pi_\theta(y_l | x)} 
\right)}_{\text{\circled{2}}} \biggr].
\end{align}
We observe the following: 
\begin{itemize}
    \item The term $\circled{1}$ shows the magnitude of the loss is proportional to $\sigma(-\rho_\theta)$.
     Thus gradient-based optimization of the DPO loss will perform stronger updates for data samples that have a low $\rho_\theta$, i.e. a low value $A_\theta$, which corresponds to \say{poor} policy performance.

    \item The term $\circled{2}$ shows the policy gradient for the chosen/rejected outputs is divided by their respective policy output probability: $\nabla_\theta\pi_\theta(y_i|x)/\pi_\theta(y_i|x)$. \emph{Thus, gradient updates for an output are enhanced if the probabilities of this output are already low.}
    This may be a problem for low-probability rejected completions 
     ($\pi_\theta(y_l|x) \ll 1$) that experience continual decrease throughout training, potentially raising the relative probability of other completions. This finding ties with the empirical observation from \cite{GeneralParadigmforPreferences} that DPO performs poorly for near-deterministic preference pairs $(\pi_\theta \in \{0, 1\})$ and requires further regularization. 
\end{itemize}

\subsection{Conclusion: An Invitation for Pessimism}
The analysis of the loss and its gradient in terms of the quantity $A_\theta$ shows DPO performs strong updates when $\circled{1}$ the target policy exudes low $A_\theta$ i.e. a low chosen/rejected likelihood ratio compared to the reference policy; or $\circled{2}$ when the probabilities $\pi_\theta(y_w|x), \pi_\theta(y_l|x)$ are both low. This is beneficial for well-labeled and abundant datasets, however, ill-labeled preference pairs likely correspond to i), and low-edit-distance and similar pairs may correspond to both i) and ii). 

This sensitivity of DPO to $\circled{1}$, and the overfitting regime $\circled{2}$ call for a mechanism to attenuate gradient updates on known weak or wrong preference pairs. If preference uncertainty scores are available, they could be leveraged as a proxy. In addition, attenuated gradient updates with a valid proxy also address an issue in DPO that all pairwise preferences are of equal weight in a dataset, despite some preference pairs being much stronger than others. Without additional attenuation, the only safety net is the quality of the reference model which regulates $A_\theta.$

\section{Method}\label{section:method}
The central contribution of this paper is to propose a penalization scheme appropriate for DPO, which leverages known uncertainty estimates on preferences. Our approach applies a reward uncertainty penalization to the overarching RLHF objective, from which we derive a new penalized DPO loss. For starters, we introduce the standard Lower Common Bound  penalisation \citep{pessimisticRL} to DPO. Our main method termed \say{Energy Factor Penalization} is a multiplicative penalization which brings considerable benefits to the binary chosen-rejected nature of DPO. 
The framework setting and notation build on that of DPO \citep{OGDPO}. In addition, we assume access to a reward model $r(x,y)$ equipped with uncertainty quantification $u(y|x)$. 

\subsection{Importing Standard Uncertainty Penalization to DPO}
Pessimistic RL subtracts a factor of the reward uncertainty $u(y|x)$ from the reward score $r(x,y)$ to obtain a conservative estimate of the reward function as a Lower Confidence Bound. Applying this to the general RLHF objective results in \Cref{eq:rlhfobjectivepenalized}.
\begin{equation}\label{eq:rlhfobjectivepenalized}
    \max_{\pi_{\theta}} \mathop{\mathbb{E}}_{\substack{x \sim D \\ y \sim \pi_{\theta}(y|x)}}
    \left[ r(x, y) -\textcolor{red}{u(y|x)} 
    \right ]
    - \beta \, D_\text{KL} \left( \pi_{\theta}(y|x) \,||\, \pi_{\text{ref}}(y|x) 
    \right). 
\end{equation}
Following prior work \citep{janpetersoperationalspacecontrol, fdivergenceminimization, OGDPO} the unique solution 
\Cref{eq:rlhfobjectivepenalized} is derived. The optimal policy $\pi^*_u$ corresponds to the reference policy being modulated by the conservative reward estimate, with $Z_u(x)$ as the appropriate partition function. Indeed, a high reward uncertainty $u(y|x)$ for a given completion $y$ will induce a lower policy probability: 
\begin{equation}\label{eq:optsolpenalized}
    \pi^*_u(y|x) = \frac{1}{Z_u(x)} \pi_{\text{ref}}(y|x) \, e^{\frac{1}{\beta} (r(x,y) -\textcolor{red}{u(y|x)})}.  
\end{equation}
Following the original DPO derivation (\Cref{app:dpoderivation}), \Cref{eq:optsolpenalized} is injected in the Bradley-Terry model, the optimal policy is replaced by the parameterized policy $\pi_\theta$ which is then optimized by maximum likelihood under the Bradley-Terry model, giving the following loss: 
\begin{align}\label{dpoobjectivepenalized}
\mathcal{L}^u_{\text{DPO}}(\pi_\theta; \pi_{\text{ref}})
&= 
\mathop{\mathbb{E}}_{(x,y_w,y_l)\sim D} 
\biggl[ -
\log 
\sigma \biggl(
\underbrace{
\hat{r}_\theta(x, y_w)- \hat{r}_\theta(x, y_l)
}_{\rho_\theta} + 
 \underbrace{\textcolor{red}{u(y_w|x)} -  \textcolor{red}{u(y_l|x)}}_{\Delta_u}
\biggr)
\biggr]
\\
\nabla_\theta \mathcal{L}^u_{\text{DPO}}
&=  
\mathop{\mathbb{E}}_{(x,y_w,y_l)\sim D}  \left[ 
- \beta
\sigma 
\left( 
- \rho_\theta -\textcolor{red}{\Delta_u}
\right)
\nabla_\theta \log \frac{\pi_\theta(y_w|x)}{\pi_\theta(y_l|x)}
\right].\label{eq:dpolossgradientpenalized}
\end{align}

\paragraph{Effect of the penalization.} The pessimistic correction amounts to adding the margin $\Delta_u = u(y_w|x) - u(y_l|x)$ between implicit rewards. Following the analysis of the loss, \Cref{eq:dpolossgradientpenalized} suggests a higher positive margin will decrease the gradient magnitude. This modification is pessimistic:
i) A high chosen reward uncertainty $u(y_w|x)$ will reduce the gradient, attenuating the gradient update, thus $\pi_\theta(y_w|x)$ will not increase much. 
ii) A high rejected reward uncertainty $u(y_l|x)$ will enhance the update: $\pi_\theta(y_l|x)$ will additionally decrease.
The chosen and rejected uncertainties in this scheme individually exhibit desirable effects, however they are not independent: they may cancel out or interfere.

\paragraph{Connection to Offset-DPO.} The derivation above recovers a DPO loss \Cref{eq:dpolossgradientpenalized} with an additional margin $\Delta_u$. This loss has the same form as Offset-DPO \citet{offsetdpo} which considers the offset $\Delta_u$ as an increasing function $f$ of the difference between reward model scores $\Delta = f(r(x,y_w)- r(x,y_l))$. They link this penalization to the Softmax-Margin loss, which we study in \Cref{app:costmargin} and derive a fully additive scheme $\Delta_u = u(y_w|x) + u(y_l|x)$ in Equation (\ref{eq:addabsolute}) which sums uncertainties, preventing their cancellation.

\subsection{Main Method: Energy Factor Penalization}
The previous section motivates a multiplicative penalization scheme (instead of subtraction) to ensure the penalization effect of either chosen or rejected uncertainties carries to the respective chosen or rejected policy gradient update terms in \Cref{eq:dpolossgradientpenalized}. Our proposed scheme multiplies the preference value or reward by an energy-like function of the uncertainty. Such penalization can be modulated by a temperature parameter $\tau > 0$:
\begin{equation}\label{eq:rlhfobjectivepenalizeddivided}
    \max_{\pi_{\theta}} \mathop{\mathbb{E}}_{\substack{x \sim D \\ y \sim \pi_{\theta}(y|x)}}
    \left[ r(x, y)\textcolor{red}{e^{-u(y|x)/\tau}} 
    \right ]
    - \beta \, D_\text{KL} \left( \pi_{\theta}(y|x) \,||\, \pi_{\text{ref}}(y|x) 
    \right). 
\end{equation}
This objective is derived into a DPO loss following the same steps as above, to obtain the expression (\ref{eq:multiplicationpenalization}). The full derivation is found in \Cref{app:derivationmultiplication}. 
\begin{align}\label{eq:multiplicationpenalization}
\mathcal{L}^u_{\text{DPO}}
&= 
\mathop{\mathbb{E}}_{(x,y_w,y_l)\sim D} 
 \biggl[ -\log \sigma 
\underbrace{\left\{ 
    \textcolor{red}{e^{u(y_w|x)/\tau}}\hat{r}_\theta(x, y_w)
    -\textcolor{red}{e^{u(y_l|x)/\tau}} \hat{r}_\theta(x, y_l)
\right\} }_{\tilde{\rho_\theta}}
 \biggr]
\\
\nabla \mathcal{L}^u_{\text{DPO}}
&=
\mathop{\mathbb{E}}_{(x,y_w,y_l)\sim D}  \left[ 
- \beta
\sigma 
\left( 
- \tilde{\rho}_\theta 
\right)
\left( 
 \textcolor{red}{e^{u(y_w|x)/\tau}}
\frac{\nabla_\theta \pi_\theta(y_w | x)}{\pi_\theta(y_w | x)} - \textcolor{red}{e^{u(y_l|x)/\tau}}\frac{\nabla_\theta \pi_\theta(y_l | x)}{\pi_\theta(y_l | x)} 
\right)
\right].\label{eq:multiplicationpenalizationgrad}
\end{align}

\paragraph{Effect of the penalization.} 
The pessimistic correction inherits similar features from the LCB scheme. However, there is an additional effect, the individual uncertainties carry onto their respective terms in the gradient of \Cref{eq:multiplicationpenalizationgrad} instead of only affecting the overall gradient magnitude. This means that uncertainties will not cancel out if they are commensurate, unlike in \Cref{eq:dpolossgradientpenalized}.
\subsection{Practical Implication: Scaling of the Penalty}

The uncertainty penalties are obtained from an external reward model, preference dataset statistics or even additional user labels. These will most likely not be to the scale of DPO's implicit rewards $\hat{r}_\theta$. Hence we apply a scalar multiplier $\alpha$ to the penalty:  $\Delta_u \leftarrow \alpha\Delta_u.$

The scaling parameter $\alpha_{z\%}$ is computed such that the penalty $\Delta_u$ is approximately to $z\%$ of the mean implicit reward ($z$ is the hyperparameter; a one standard deviation penalty of the reward roughly corresponds to $z =30\%$). The same principle is applied to the temperature parameter $\tau_{z\%}$ for multiplication penalty. Denote the mean implicit reward as $\bar{r}_{\theta}$, and the mean uncertainty value $\bar{u}$, the scaling parameter is computed as follows:
\begin{align}
    \alpha_z \, \Delta_u &= (1-z) \, \bar{r}_{\theta} \implies \alpha_z = (1-z) \, \bar{r}_{\theta}\,  / \, {\Delta_u} \\
   e^{\bar{u}/\tau_z} &= (1+z) \quad \implies\,  \tau_z = \bar{u} \, /\,  \log(1+z) 
\end{align}
Naturally, implicit reward values evolve throughout training which motivates the use of an exponential moving average estimate of the mean reward. In practice, we compute this every batch of training: 
$
\bar{r}_{\theta}^{(t)} = \lambda \bar{r}_{\theta}^{(t-1)} + (1-\lambda) \hat{r}_{\theta}^{(t)},
$
where \(\bar{r}_{\theta}^{(t)}\) and \(\bar{r}_{\theta}^{(t-1)}\) denote the moving average estimates at batch $t$ and $t-1$ respectively, $\bar{r}_{\theta}^{(t)}$ denotes the mean implicit reward of batch $t$, and \(\lambda\in [0,1)\) is the decay factor which controls the influence of previous estimates, balancing between smoothness and responsiveness of the moving average. The same is applied to estimate $\bar{u}$ in the case of multiplication penalty.

\subsection{Generalization to $\Psi$PO and IPO}
The $\Psi$PO framework by \citet{GeneralParadigmforPreferences} is a general objective that encompasses many RLHF methods, including the standard RLHF objective (\ref{eq:rlhfobjective}), DPO, and Identity Preference Optimization (IPO). We generalize our penalization schemes to $\Psi$PO in \Cref{app:generelisationPsi}, and import our schemes to IPO. Our schemes for IPO are evaluated empirically.

\subsection{Summary of Proposed Penalizations.} 
We present the initial LCB addition penalization, our main method (multiplication) and other uncertainty penalization schemes. The Cost-Margin-motivated penalizations are derived in \Cref{app:costmargin} and presented as ablations over different ways to include uncertainty in DPO. The reward model free penalizations are not empirically evaluated and are dedicated to future work. 
\begin{table}[H]
\centering
\renewcommand{\arraystretch}{1} 
\begin{tabular}{@{}llll@{}}
\toprule
Type & Motivation & Name & Margin or Modification \\ \midrule
\multirow[t]{4}{*}{Reward Model Based} & 
LCB
&
Addition & $\Delta_u = u(y_w|x) - u(y_l|x) $
\\[5pt]
 & LCB & Multiplication & 
 $\hat{r}_\theta(y|x) \leftarrow e^{u(y|x)/\tau}\, \hat{r}_\theta(y|x) $
 \\[5pt]
 & Cost Margin & Addition Absolute & $\Delta_u = |u(y_w|x) + u(y_l|x) |$
 \\[5pt]
 & Cost Margin & Probability &  $\Delta_u = \Phi \left ( \frac{\bar{r}_l - \bar{r}_w}{ \sqrt{u_l^2 + u_w^2}}
    \right)$
 \\ \\
\multirow[t]{2}{*}{Reward Model Free} & 
Cost Margin
& Predictive Entropy & $\Delta_u = \frac{1}{N} \sum_{n=1}^{N} \log \pi_{\theta}(y^{n}|x)$
\\ 
&&& or \\
&&&
$
\Delta_u = \sigma \left(\frac{1}{N} \sum_{n=1}^{N} \log \pi_{\theta}(y^{n}|x) - B
\right) $
 \\[5pt] \bottomrule
\end{tabular}%
\caption{Proposed Uncertainty Penalization Schemes.}
\label{tab:proposedschemes}
\end{table}
\vspace{-1cm}

\section{Experiments}

\subsection{Experimental Setup}

An ensemble of reward models is trained to compute uncertainty estimates for the Anthropic-HH dataset. All LLMs first undergo supervised fine-tuning (SFT) on the chosen completions of the dataset, and then preference fine-tuning is performed. The evaluation is done by scoring model completions for prompts from the Anthropic-HH test set. The experiments are performed on GPT2 Medium (355M weights, pretrained); complete implementation details and hyperparameter search ranges are found in \Cref{app:experimentdetails}.\footnote{The code is available at: \href{https://anonymous.4open.science/r/PessimisticDPO-DAF4/}{https://anonymous.4open.science/r/PessimisticDPO-DAF4/}}
\paragraph{Dataset.} The Anthropic-HH dataset \citep{anthropicdata} consists of 160'800 train and 8552 test records of chosen and a rejected human-assistant conversations.
\paragraph{Reward Model Ensemble.} 5 individual reward models are trained on shuffled 90\% splits of the Anthropic-HH training dataset. GPT2 is used with a regression head, and trained via the Huggingface TRL library's RewardTrainer with default arguments (1 epoch). The ensemble obtains a mean classification accuracy of 67\% on the test dataset. 
\paragraph{SFT Reference.} SFT training is performed in completion-only mode on chosen completions using the TRL SFT Trainer with a linearly decreasing learning rate of \texttt{$1.45e^{-5}$}, 8 batch size, 8 gradient accumulation steps, 10\% warmup for 1 epoch, and no LoRA. 
\paragraph{DPO Models and Baseline.} Fine-tuned models were trained on top of the SFT reference using LoRA, with an initial hyperparameter search. For the DPO baseline with GPT2 Medium, the optimal parameters were $\beta=0.6$, \texttt{1e-7} learning rate with linear decrease schedule,  $32$ batch size, no gradient accumulation, LoRA parameters $(r,\alpha)=(16,16)$, and 10\% warmup for 1 epoch. Pessimistic DPO ablations used the same parameters. 
\subsection{Performance Evaluation}
\begin{figure}[h]
     \centering
     \begin{subfigure}[c]{0.4\linewidth}
         \centering
         \includegraphics[width=\linewidth]{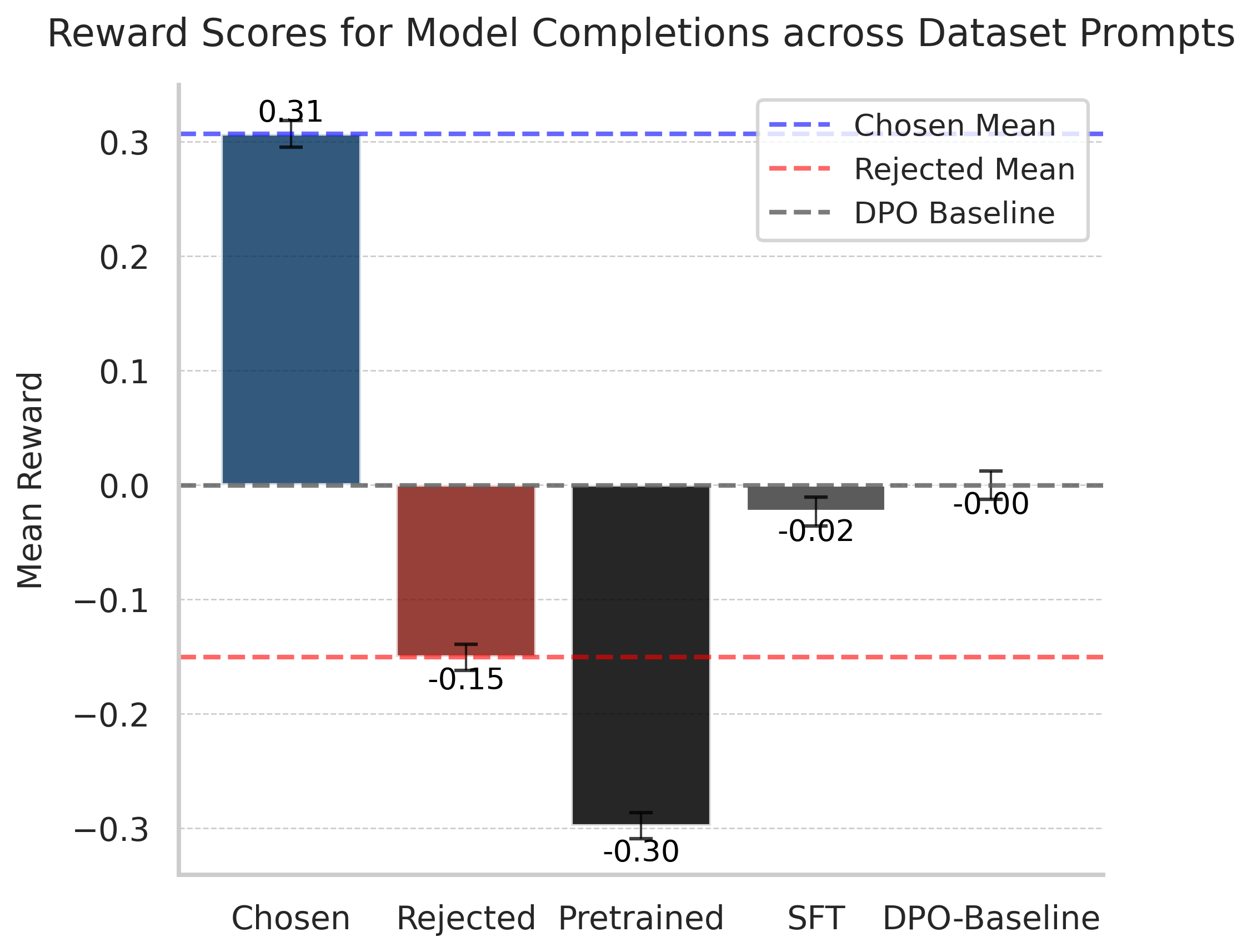}
         \caption{Experimental Setup. The increase in scores from pretrained to DPO validates the training setup.}
         \label{fig:rewardsdistr}
     \end{subfigure}
     \hfill
     \begin{subfigure}[c]{0.57\linewidth}
         \centering
         \includegraphics[width=\linewidth]{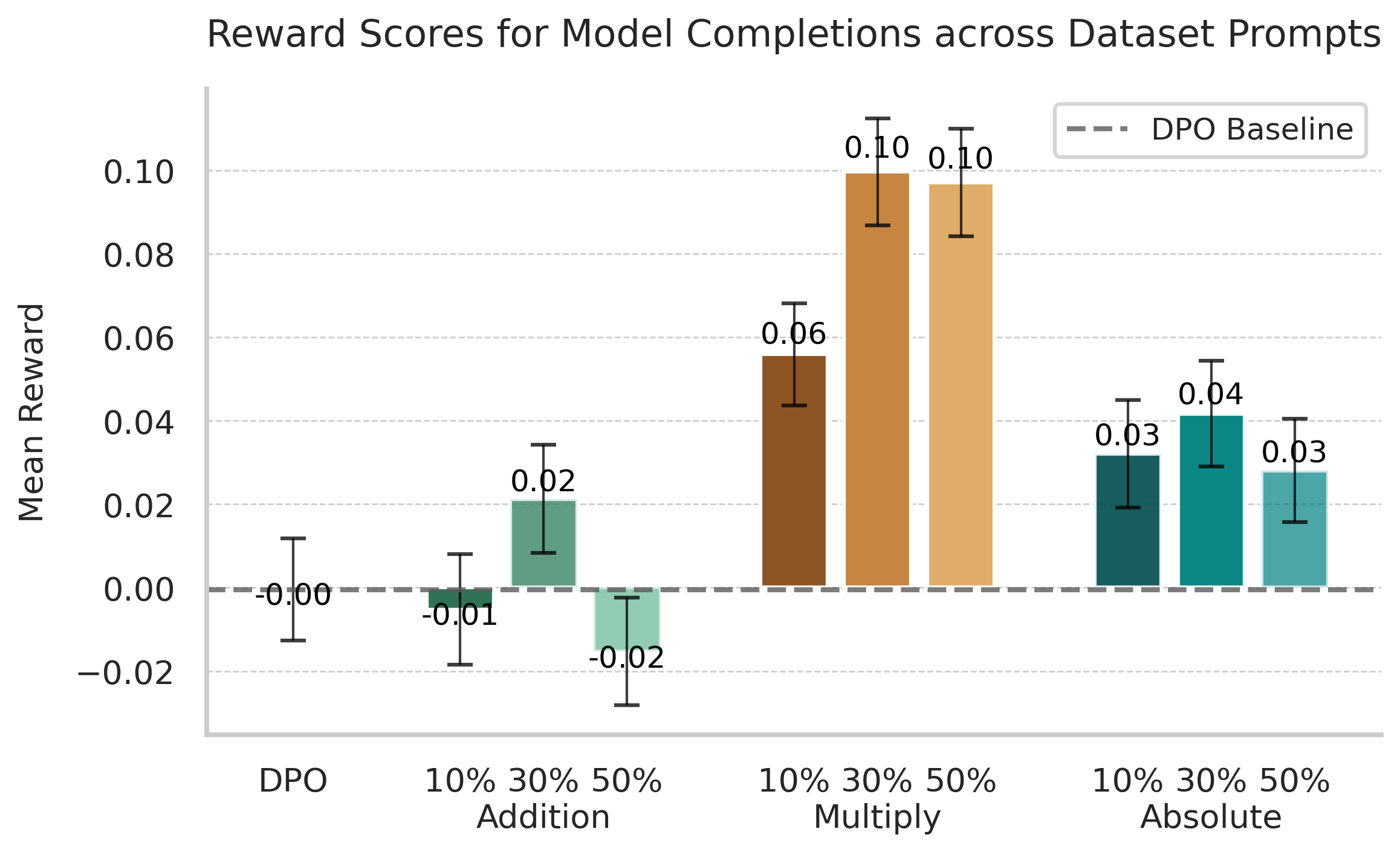}
         \caption{Evaluation of Finetunings. Pessimistic schemes perform on-par or better than DPO, best scores achieved at 30\% penalty.}
         \label{fig:dpobarchart}
     \end{subfigure}
        \vspace{-0.3cm}
        \caption{Rewards over completions for 500 Anthropic-HH test prompts.}
    \label{fig:rewardssss}
    \vspace{-0.55cm}
\end{figure}

The penalized DPO models perform on par or better than vanilla DPO. Figure \ref{fig:dpobarchart} shows the addition scheme performs similarly to DPO whereas our multiplication scheme outclasses the baseline for all penalization strengths. For each of the schemes, the middle penalization strength of $30\%$ performs best. Scores and uncertainties across all models and chosen/rejected baselines, are presented in table \ref{tab:demresults}, showing our schemes outperform the SFT and DPO baselines.

\subsection{Robustness Evaluation}

\begin{figure}[h]
     \centering
     \begin{subfigure}[b]{0.58\textwidth}
         \includegraphics[width=\textwidth]{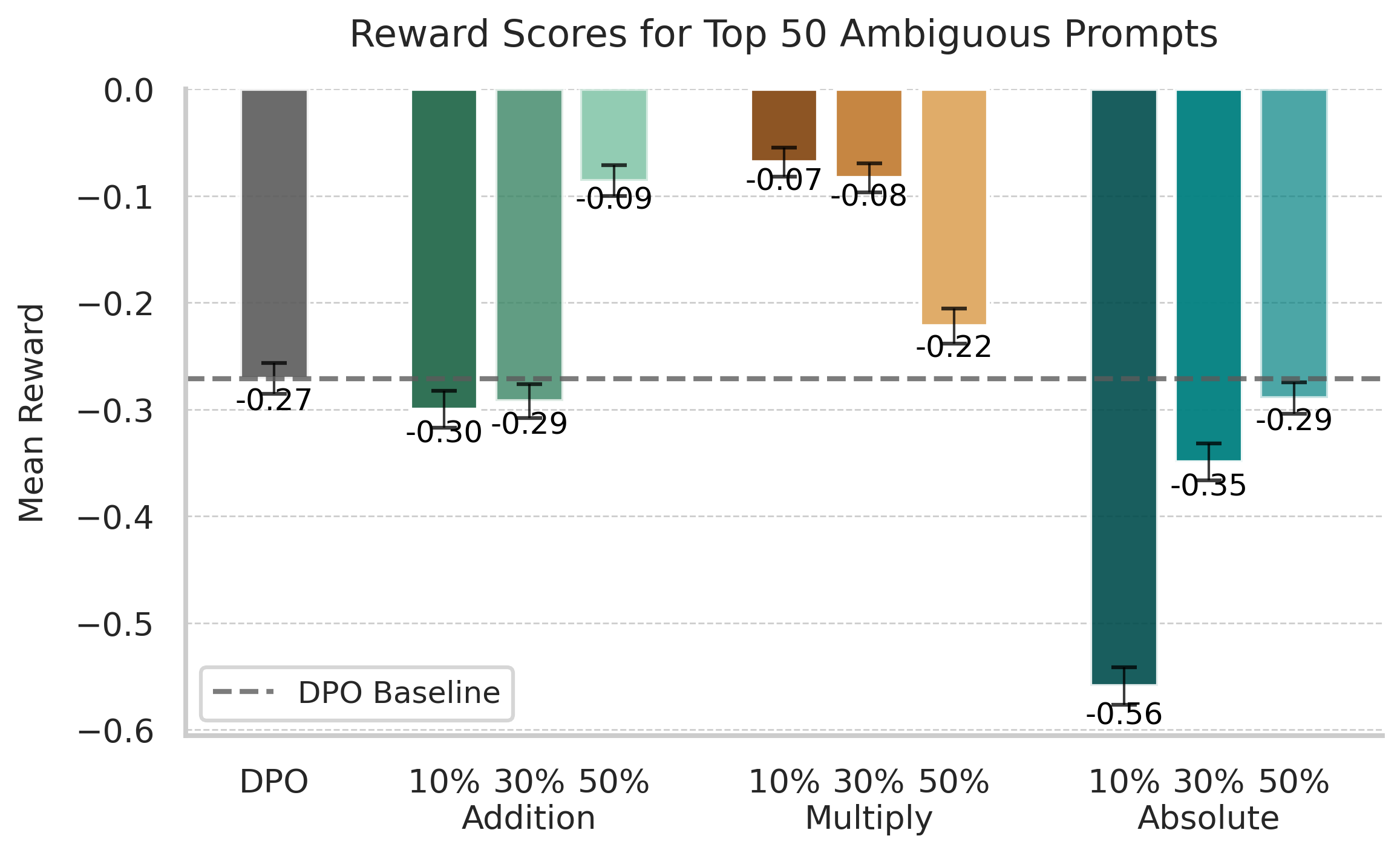}
         \caption{Evaluation on 50 prompts with highest chosen/rejected text uncertainty. Multiplication penalty perform best. Scores improve with penalization strength for Addition and Absolute.}
         \label{fig:ambigbar}  
     \end{subfigure}
     \hfill
     \begin{subfigure}[b]{0.4\textwidth}
         \includegraphics[width=\textwidth]{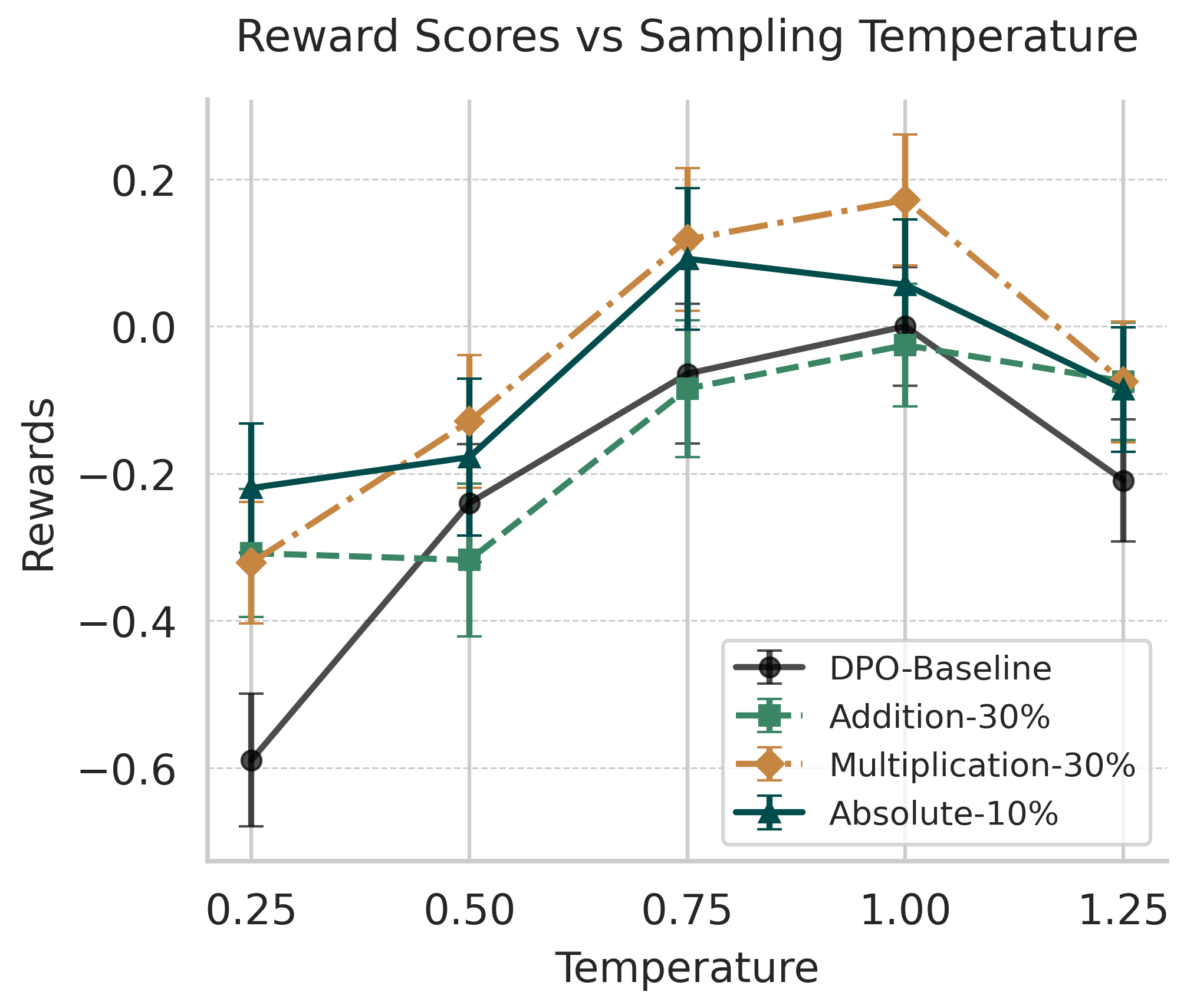}
         \caption{Mean reward for 200 completions per temperature. Vanilla DPO rewards drop the most as temperature varies.}
         \label{fig:temps}
     \end{subfigure}
     \vspace{-0.1cm}
     \caption{Study of robustness to reward overoptimization.}
     \label{fig:robustness}
     \vspace{-0.3cm}
\end{figure}

\paragraph{Performance on Uncertain Samples.} Our method should improve training on chosen and rejected pairs exhibiting high reward uncertainty. To evaluate this, we isolate the 50 test records whose chosen and rejected responses have the highest summed reward uncertainty. \Cref{fig:ambigbar} depicts reward scores for tuned model completions on those 50 prompts and shows superior performance of the multiplication scheme. Addition performs similarly to the DPO baseline while absolute scheme performs worse. For both addition and absolute, performance increases with penalty strength. 

\paragraph{Study of Sampling Temperature.} 
\Cref{fig:temps} displays mean reward scores for generated completions at different sampling temperatures from the top-performing models of each scheme. The multiplication and absolute schemes perform better than DPO across all temperatures. All curves are bell-shaped: low temperatures favour high-probability, repetitive text, and high temperatures invite randomness in next-token sampling leading to to more creative text at the risk of incoherence or quality. For well-behaved policies we expect the scores to be more level at various temperatures. The penalized models indeed have a lower drop-off in performance across the temperature range.

\definecolor{mygold}{RGB}{218,165,32}
\definecolor{mybronze}{RGB}{160,82,45}
\definecolor{mysilver}{RGB}{65, 65, 65}

    \begin{minipage}[]{0.5\linewidth}
        \resizebox{\linewidth}{!}{
        \begin{tabular}{@{}lrr@{}}
        \toprule
        \multicolumn{1}{l}{\multirow{2}{*}{\textbf{Model}}} & \multicolumn{1}{c}{\textbf{All Prompts}} & \multicolumn{1}{c}{\textbf{Top 50 Ambiguous}} \\
         & \multicolumn{1}{c}{Mean reward} & \multicolumn{1}{c}{Mean reward} \\
        \midrule
        Chosen & $0.306 \pm 0.013$ & $-0.350 \pm 0.018$ \\[2pt] 
        Rejected & $-0.151 \pm 0.013$ & $-0.424 \pm 0.017$ \\ \midrule 
        Pretrained & $-0.298 \pm 0.009$ & $-0.524 \pm 0.011$ \\[2pt] 
        SFT & $-0.026 \pm 0.011$ & $-0.253 \pm 0.013$ \\[2pt] 
        DPO & $-0.001 \pm 0.011$ & $-0.271 \pm 0.012$ \\[2pt] 
        Addition (10\%) & $-0.005 \pm 0.012$ & $-0.300 \pm 0.013$ \\[2pt] 
        Addition (30\%) & $0.021 \pm 0.011$ & $-0.292 \pm 0.014$ \\[2pt] 
        Addition (50\%) & $-0.015 \pm 0.011$ & \textcolor{mybronze}{$\mathbf{-0.086 \pm 0.012}$} \\[2pt] 
        Multiplication (10\%) & \textcolor{mybronze}{$\mathbf{0.056 \pm 0.011}$} & \textcolor{mygold}{$\mathbf{-0.069 \pm 0.014}$} \\[2pt] 
        Multiplication (30\%) & \textcolor{mygold}{$\mathbf{0.099 \pm 0.011}$} & \textcolor{mysilver}{$\mathbf{-0.083 \pm 0.012}$} \\[2pt] 
        Multiplication (50\%) & \textcolor{mysilver}{$\mathbf{0.097 \pm 0.011}$} & $-0.222 \pm 0.014$ \\[2pt] 
        Absolute (10\%) & $0.032 \pm 0.011$ & $-0.559 \pm 0.013$ \\[2pt] 
        Absolute (30\%) & $0.042 \pm 0.011$ & $-0.349 \pm 0.015$ \\[2pt] 
        Absolute (50\%) & $0.028 \pm 0.011$ & $-0.289 \pm 0.014$ \\
        \bottomrule
        \end{tabular}%
        }
        \captionof{table}{Reward for Model Completions on 500 Anthropic-HH Test Set prompts. Values are presented as the mean $\pm$ standard error. Top three scores are highlighted in \textcolor{mygold}{\textbf{gold}}, \textcolor{mysilver}{\textbf{silver}} and \textcolor{mybronze}{\textbf{bronze}}. }
        \label{tab:demresults}

    \end{minipage}%
    \hfill
    \begin{minipage}[]{0.45\textwidth}   
        \paragraph{Discussion.} The reward model ensemble provides a reasonable assessment of completions relative to chosen and rejected outputs, as it obtained a 67\% test accuracy. However, its reward margin is small (Figures \ref{fig:rewardsdistr}, \ref{fig:rewardssss}) and may overfit to Anthropic-style responses, repercussing to inaccurate rewards for different but still preferable text; this may be addressed by a more powerful model.
        \bigbreak 
        Example completions in \Cref{app:completions} show the fine-tuned models exhibit agreeableness and coherence. However, training the DPO Baseline to a satisfactory performance (\Cref{fig:rewardsdistr}) required extensive hyperparameter tuning. This may be due to the hardness of learning preference associations in Anthropic-HH relative to GPT2 Medium’s size. Despite this, the results are promising for larger models and diverse tasks.
    \end{minipage}

\section{Conclusion}
This work proposes a new framework for DPO inspired by pessimistic offline RL that integrates pessimism into DPO by leveraging preference uncertainty estimates. The derived penalization schemes are tailored to the binary nature of DPO, with our Energy Factor scheme performing best overall and robustness-wise in our illustrative experiments. The empirical findings invite further evaluation with more powerful models on various tasks (summarization, dialogue, completion...). Finally, a generalization to $\Psi$PO, IPO, and a reward-model-free scheme are proposed; further work is invited to develop and implement these.

\newpage 
\bibliography{iclr2025_conference}

\begin{thebibliography}{25}
\providecommand{\natexlab}[1]{#1}
\providecommand{\url}[1]{\texttt{#1}}
\expandafter\ifx\csname urlstyle\endcsname\relax
  \providecommand{\doi}[1]{doi: #1}\else
  \providecommand{\doi}{doi: \begingroup \urlstyle{rm}\Url}\fi

\bibitem[A. \& Terry(1952)A. and Terry]{bradleyT}
Bradley~R. A. and M.~E. Terry.
\newblock Rank analysis of incomplete block designs: I. the method of paired comparisons.
\newblock \emph{Biometrika, 39(3/4):324–345}, 1952.

\bibitem[Amini et~al.(2024)Amini, Vieira, and Cotterell]{offsetdpo}
Afra Amini, Tim Vieira, and Ryan Cotterell.
\newblock Direct preference optimization with an offset, 2024.

\bibitem[Azar et~al.(2023)Azar, Rowland, Piot, Guo, Calandriello, Valko, and Munos]{GeneralParadigmforPreferences}
Mohammad~Gheshlaghi Azar, Mark Rowland, Bilal Piot, Daniel Guo, Daniele Calandriello, Michal Valko, and Rémi Munos.
\newblock A general theoretical paradigm to understand learning from human preferences, 2023.

\bibitem[Bai et~al.(2022)Bai, Jones, Ndousse, Askell, Chen, DasSarma, Drain, Fort, Ganguli, Henighan, Joseph, Kadavath, Kernion, Conerly, El-Showk, Elhage, Hatfield-Dodds, Hernandez, Hume, Johnston, Kravec, Lovitt, Nanda, Olsson, Amodei, Brown, Clark, McCandlish, Olah, Mann, and Kaplan]{anthropicdata}
Yuntao Bai, Andy Jones, Kamal Ndousse, Amanda Askell, Anna Chen, Nova DasSarma, Dawn Drain, Stanislav Fort, Deep Ganguli, Tom Henighan, Nicholas Joseph, Saurav Kadavath, Jackson Kernion, Tom Conerly, Sheer El-Showk, Nelson Elhage, Zac Hatfield-Dodds, Danny Hernandez, Tristan Hume, Scott Johnston, Shauna Kravec, Liane Lovitt, Neel Nanda, Catherine Olsson, Dario Amodei, Tom Brown, Jack Clark, Sam McCandlish, Chris Olah, Ben Mann, and Jared Kaplan.
\newblock Training a helpful and harmless assistant with reinforcement learning from human feedback, 2022.

\bibitem[Casper et~al.(2023)Casper, Davies, Shi, Gilbert, Scheurer, Rando, Freedman, Korbak, Lindner, Freire, Wang, Marks, Segerie, Carroll, Peng, Christoffersen, Damani, Slocum, Anwar, Siththaranjan, Nadeau, Michaud, Pfau, Krasheninnikov, Chen, Langosco, Hase, Bıyık, Dragan, Krueger, Sadigh, and Hadfield-Menell]{openproblemsinRLHF}
Stephen Casper, Xander Davies, Claudia Shi, Thomas~Krendl Gilbert, Jérémy Scheurer, Javier Rando, Rachel Freedman, Tomasz Korbak, David Lindner, Pedro Freire, Tony Wang, Samuel Marks, Charbel-Raphaël Segerie, Micah Carroll, Andi Peng, Phillip Christoffersen, Mehul Damani, Stewart Slocum, Usman Anwar, Anand Siththaranjan, Max Nadeau, Eric~J. Michaud, Jacob Pfau, Dmitrii Krasheninnikov, Xin Chen, Lauro Langosco, Peter Hase, Erdem Bıyık, Anca Dragan, David Krueger, Dorsa Sadigh, and Dylan Hadfield-Menell.
\newblock Open problems and fundamental limitations of reinforcement learning from human feedback, 2023.

\bibitem[Gimpel \& Smith(2010)Gimpel and Smith]{softmaxmargin}
Kevin Gimpel and Noah~A. Smith.
\newblock Softmax-margin {CRF}s: Training log-linear models with cost functions.
\newblock In Ron Kaplan, Jill Burstein, Mary Harper, and Gerald Penn (eds.), \emph{Human Language Technologies: The 2010 Annual Conference of the North {A}merican Chapter of the Association for Computational Linguistics}, June 2010.

\bibitem[Go et~al.(2022)Go, Korbak, Kruszewski, Rozen, Ryu, and Dymetman]{fdivergenceminimization}
Dongyoung Go, Tomasz Korbak, Germán Kruszewski, Jos Rozen, Nahyeon Ryu, and Marc Dymetman.
\newblock Aligning language models with preferences through f-divergence minimization, 2022.

\bibitem[Hu et~al.(2021)Hu, Shen, Wallis, Allen-Zhu, Li, Wang, Wang, and Chen]{loraboy}
Edward~J. Hu, Yelong Shen, Phillip Wallis, Zeyuan Allen-Zhu, Yuanzhi Li, Shean Wang, Lu~Wang, and Weizhu Chen.
\newblock Lora: Low-rank adaptation of large language models, 2021.

\bibitem[Huang et~al.(2024)Huang, Zhan, Xie, Lee, Sun, Krishnamurthy, and Foster]{chisquared}
Audrey Huang, Wenhao Zhan, Tengyang Xie, Jason~D. Lee, Wen Sun, Akshay Krishnamurthy, and Dylan~J. Foster.
\newblock Correcting the mythos of kl-regularization: Direct alignment without overoptimization via chi-squared preference optimization, 2024.

\bibitem[Jin et~al.(2020)Jin, Yang, and Wang]{pessimisticRL}
Ying Jin, Zhuoran Yang, and Zhaoran Wang.
\newblock Is pessimism provably efficient for offline rl?
\newblock \emph{CoRR}, abs/2012.15085, 2020.

\bibitem[Kadavath et~al.(2022{\natexlab{a}})Kadavath, Conerly, Askell, Henighan, Drain, Perez, Schiefer, Hatfield-Dodds, DasSarma, Tran-Johnson, Johnston, El-Showk, Jones, Elhage, Hume, Chen, Bai, Bowman, Fort, Ganguli, Hernandez, Jacobson, Kernion, Kravec, Lovitt, Ndousse, Olsson, Ringer, Amodei, Brown, Clark, Joseph, Mann, McCandlish, Olah, and Kaplan]{predictiveentropy}
Saurav Kadavath, Tom Conerly, Amanda Askell, Tom Henighan, Dawn Drain, Ethan Perez, Nicholas Schiefer, Zac Hatfield-Dodds, Nova DasSarma, Eli Tran-Johnson, Scott Johnston, Sheer El-Showk, Andy Jones, Nelson Elhage, Tristan Hume, Anna Chen, Yuntao Bai, Sam Bowman, Stanislav Fort, Deep Ganguli, Danny Hernandez, Josh Jacobson, Jackson Kernion, Shauna Kravec, Liane Lovitt, Kamal Ndousse, Catherine Olsson, Sam Ringer, Dario Amodei, Tom Brown, Jack Clark, Nicholas Joseph, Ben Mann, Sam McCandlish, Chris Olah, and Jared Kaplan.
\newblock Language models (mostly) know what they know, 2022{\natexlab{a}}.

\bibitem[Kadavath et~al.(2022{\natexlab{b}})Kadavath, Conerly, Askell, Henighan, Drain, Perez, Schiefer, Hatfield-Dodds, DasSarma, Tran-Johnson, Johnston, El-Showk, Jones, Elhage, Hume, Chen, Bai, Bowman, Fort, Ganguli, Hernandez, Jacobson, Kernion, Kravec, Lovitt, Ndousse, Olsson, Ringer, Amodei, Brown, Clark, Joseph, Mann, McCandlish, Olah, and Kaplan]{entropyerror}
Saurav Kadavath, Tom Conerly, Amanda Askell, Tom Henighan, Dawn Drain, Ethan Perez, Nicholas Schiefer, Zac Hatfield-Dodds, Nova DasSarma, Eli Tran-Johnson, Scott Johnston, Sheer El-Showk, Andy Jones, Nelson Elhage, Tristan Hume, Anna Chen, Yuntao Bai, Sam Bowman, Stanislav Fort, Deep Ganguli, Danny Hernandez, Josh Jacobson, Jackson Kernion, Shauna Kravec, Liane Lovitt, Kamal Ndousse, Catherine Olsson, Sam Ringer, Dario Amodei, Tom Brown, Jack Clark, Nicholas Joseph, Ben Mann, Sam McCandlish, Chris Olah, and Jared Kaplan.
\newblock Language models (mostly) know what they know, 2022{\natexlab{b}}.

\bibitem[Li et~al.(2021)Li, Tang, Tomizuka, and Zhan]{pessimisticRLdealingwiththeunknwown}
Jinning Li, Chen Tang, Masayoshi Tomizuka, and Wei Zhan.
\newblock Dealing with the unknown: Pessimistic offline reinforcement learning.
\newblock In \emph{5th Annual Conference on Robot Learning}, 2021.

\bibitem[Morimura et~al.(2024)Morimura, Sakamoto, Jinnai, Abe, and Ariu]{filtereddpo}
Tetsuro Morimura, Mitsuki Sakamoto, Yuu Jinnai, Kenshi Abe, and Kaito Ariu.
\newblock Filtered direct preference optimization, 2024.

\bibitem[Muldrew et~al.(2024)Muldrew, Hayes, Zhang, and Barber]{activedpo}
William Muldrew, Peter Hayes, Mingtian Zhang, and David Barber.
\newblock Active preference learning for large language models, 2024.

\bibitem[Pal et~al.(2024)Pal, Karkhanis, Dooley, Roberts, Naidu, and White]{DPOP}
Arka Pal, Deep Karkhanis, Samuel Dooley, Manley Roberts, Siddartha Naidu, and Colin White.
\newblock Smaug: Fixing failure modes of preference optimisation with dpo-positive, 2024.

\bibitem[Peters \& Schaal(2007)Peters and Schaal]{janpetersoperationalspacecontrol}
Jan Peters and Stefan Schaal.
\newblock Reinforcement learning by reward-weighted regression for operational space control.
\newblock In \emph{Proceedings of the 24th International Conference on Machine Learning}, 2007.

\bibitem[Rafailov et~al.(2023)Rafailov, Sharma, Mitchell, Manning, Ermon, and Finn]{OGDPO}
Rafael Rafailov, Archit Sharma, Eric Mitchell, Christopher~D Manning, Stefano Ermon, and Chelsea Finn.
\newblock Direct preference optimization: Your language model is secretly a reward model.
\newblock In \emph{Thirty-seventh Conference on Neural Information Processing Systems}, 2023.

\bibitem[Shen et~al.(2024)Shen, Daheim, Cong, Nickl, Marconi, Bazan, Yokota, Gurevych, Cremers, Khan, and Möllenhoff]{variationaleffective}
Yuesong Shen, Nico Daheim, Bai Cong, Peter Nickl, Gian~Maria Marconi, Clement Bazan, Rio Yokota, Iryna Gurevych, Daniel Cremers, Mohammad~Emtiyaz Khan, and Thomas Möllenhoff.
\newblock Variational learning is effective for large deep networks, 2024.

\bibitem[Xiong et~al.(2024)Xiong, Dong, Ye, Wang, Zhong, Ji, Jiang, and Zhang]{iterativeDPO}
Wei Xiong, Hanze Dong, Chenlu Ye, Ziqi Wang, Han Zhong, Heng Ji, Nan Jiang, and Tong Zhang.
\newblock Iterative preference learning from human feedback: Bridging theory and practice for rlhf under kl-constraint, 2024.

\bibitem[Xu et~al.(2024)Xu, Fu, Gao, Ye, Liu, Mei, Wang, Yu, and Wu]{dposuperiorppo}
Shusheng Xu, Wei Fu, Jiaxuan Gao, Wenjie Ye, Weilin Liu, Zhiyu Mei, Guangju Wang, Chao Yu, and Yi~Wu.
\newblock Is dpo superior to ppo for llm alignment? a comprehensive study, 2024.

\bibitem[Yang et~al.(2024{\natexlab{a}})Yang, Robeyns, Coste, Wang, Bou-Ammar, and Aitchison]{bayesianreward}
Adam~X. Yang, Maxime Robeyns, Thomas Coste, Jun Wang, Haitham Bou-Ammar, and Laurence Aitchison.
\newblock Bayesian reward models for llm alignment, 2024{\natexlab{a}}.

\bibitem[Yang et~al.(2024{\natexlab{b}})Yang, Robeyns, Coste, Wang, Bou-Ammar, and Aitchison]{bayesianrwdmodel}
Adam~X. Yang, Maxime Robeyns, Thomas Coste, Jun Wang, Haitham Bou-Ammar, and Laurence Aitchison.
\newblock Bayesian reward models for llm alignment, 2024{\natexlab{b}}.

\bibitem[Zhai et~al.(2023{\natexlab{a}})Zhai, Zhang, Lei, Yu, Xu, Feng, Ding, and Wang]{LORARLHF}
Yuanzhao Zhai, Han Zhang, Yu~Lei, Yue Yu, Kele Xu, Dawei Feng, Bo~Ding, and Huaimin Wang.
\newblock Uncertainty-penalized reinforcement learning from human feedback with diverse reward lora ensembles, 2023{\natexlab{a}}.

\bibitem[Zhai et~al.(2023{\natexlab{b}})Zhai, Zhang, Lei, Yu, Xu, Feng, Ding, and Wang]{diverselora}
Yuanzhao Zhai, Han Zhang, Yu~Lei, Yue Yu, Kele Xu, Dawei Feng, Bo~Ding, and Huaimin Wang.
\newblock Uncertainty-penalized reinforcement learning from human feedback with diverse reward lora ensembles, 2023{\natexlab{b}}.

\end{thebibliography}
\bibliographystyle{iclr2025_conference}
\newpage 
\appendix
\section*{Appendix}
\section{DPO Background and Derivation}\label{app:dpoderivation}

\paragraph{Problem setup.}
We have a static dataset of comparisons denoted as $\mathcal{D} = \{x, y_w, y_l\}_{i=1}^{N}$, which is usually obtained by prompting an SFT (supervised-fine-tuned) model with prompts $x$ to produce pairs of answers $(y_w, y_l)$ (these comparisons do not have scores, they are absolute). Human labellers identify the preferred output $y_w$ over the undesired $y_l$. 
We define a reference (SFT) model policy as $\pi_{\text{ref}}$ and our parameterized policy as $\pi_\theta$, which we aim to fit as to make the preferred outputs $y_w$ more likely, while staying close to the reference policy.

\subsection{DPO Derivation from RLHF.} The RLHF objective is: 

\begin{equation}\label{eq:rlhfobjective}
    \max_{\pi_{\theta}} \mathop{\mathbb{E}}_{\substack{x \sim D \\ y \sim \pi_{\theta}(y|x)}}
    \left[ r_{\phi}(x, y)
    \right ]
    - \beta \, \text{KL} \left( \pi_{\theta}(y|x) \,||\, \pi_{\text{ref}}(y|x) 
    \right). 
\end{equation}
By factoring the terms under the expectation, we obtain an KL divergence-like expression, we introduce the partition function\footnote{Some works show this function is very close to 1.}  $Z(x) = \sum_{y} \pi_{\text{ref}}(y | x) \exp\left(\frac{1}{\beta} r(x, y)\right)$ to normalize the denominator, and the optimal value annuls this KL-divergence, giving us the unique optimal solution: 
\begin{equation}\label{eq:optsolappendix}
    \pi^*(y|x) = \frac{1}{Z(x)} \pi_{\text{ref}}(y|x) \, e^{\frac{1}{\beta}r(x,y)}.  
\end{equation}
Intuitively, our optimal policy aligns with the reference policy, modulated by high or low rewards of certain outputs. Now, equation \ref{eq:optsolappendix} can be re-arranged to express the ground-truth reward model in function of the induced optimal and reference policies: $r(x, y) = \beta \log \left( \frac{\pi^*(y | x)}{\pi_{\text{ref}}(y | x)} \right) + \beta \log Z(x)$.

The authors, define the $\hat{r}_\theta(x, y) = \beta \log \left( \frac{\pi_\theta(y | x)}{\pi_{\text{ref}}(y | x)} \right) + \beta \log Z(x) \approx \beta \log \left( \frac{\pi_\theta(y | x)}{\pi_{\text{ref}}(y | x)} \right)$, as the reward implicitly defined by the language model $\pi_\theta.$ 

Enter, the Bradley-Terry model: $p(y_1 \succ y_2 | x) = \sigma(r(x, y_1) - r(x, y_2))$, interpreted as the probability of answer $y_1$ being favoured over $y_2$ as a function of their human-labelled rewards. In RLHF the reward model is trained to maximize $p(y_w, y_l)$ over a dataset. For DPO, we do the same, by substituting the parameterised reward expressions: 
\begin{align}\label{dpoobjectiveappendix}
\mathcal{L}_{\text{DPO}}(\pi_\theta; \pi_{\text{ref}})
&= 
-
\mathop{\mathbb{E}}_{(x,y_w,y_l)\sim D} 
\left[ 
\log 
p(y_w \succ y_l)
\right]
\\
&=
-\mathop{\mathbb{E}}_{(x,y_w,y_l)\sim D} 
\left[ 
\log 
\sigma \left (
\underbrace{
\hat{r}_\theta(x, y_w)- \hat{r}_\theta(x, y_l) 
}_{\rho_\theta}
\right)
\right]
\\
&=
-\mathop{\mathbb{E}}_{(x,y_w,y_l)\sim D} \left[ \log \sigma \left( 
\beta \log \frac{\pi_\theta(y_w | x)}{\pi_{\text{ref}}(y_w | x)} - \beta \log \frac{\pi_\theta(y_l | x)}{\pi_{\text{ref}}(y_l | x)} 
\right) \right].
\end{align}

\subsection{DPO as a Binary Classification Problem.}
Another view interprets the DPO loss \ref{eq:dpoobjective} as akin to binary classification, where the given $y_1$ is preferable to $y_2$, and we aim to train the parametrized policy $\pi_\theta$ such that the preference model $p_\theta(y_1 \succ y_2 | x) = \sigma(\hat{r}_\theta(x, y_1) - \hat{r}_\theta(x, y_2))$ predicts 1, meaning we aim to maximize the quantity $\rho_\theta$ (see GPO paper for proper derivation), which effectively increases the margin between the probability of the preferred and unpreferred sample. Note: sometimes in DPO optimization, both probabilities are decreased, but the unpreferred sample probability is decreased more strongly; while this improves performance on the training preference dataset this might have the adverse effect of increasing the probability of other output text sequences.

\subsection{DPO Loss Gradient Derivation}
We provide the full derivation for the loss gradient w.r.t. policy parameters. Recall the properties of the sigmoid $\sigma'(x) = \sigma(x) (1-\sigma(x))$, $\sigma(-x) = 1-\sigma(x)$, with $\nabla \log(x) = \frac{\nabla x}{x}$.
\begin{align}\label{eq:dpolossappendix}
\nabla_\theta \, \ell_{\text{DPO}}(x, y_w, y_l; \theta)
&= - \nabla_\theta \log \sigma \left( \rho_\theta \right) \\
&= - \frac{\nabla_\theta \sigma \left( \rho_\theta \right)}{\sigma \left( \rho_\theta \right) }\\
&= - \frac{\sigma(\rho_\theta)(1- \sigma(\rho_\theta)) \, \nabla_\theta \rho_\theta}
{\sigma(\rho_\theta)} \\
&= - \sigma(- \rho_\theta)  \, \nabla_\theta \rho_\theta \\
&=  - \beta \sigma(- \rho_\theta) \left (
\frac{\nabla_\theta \pi_\theta(y_w | x)}{\pi_\theta(y_w | x)} -\frac{\nabla_\theta \pi_\theta(y_w | x)}{\pi_\theta(y_w | x)} 
\right) \\
&=
- \beta \sigma(- \rho_\theta)
\left( 
\frac{\nabla_\theta \pi_\theta(y_w | x)}{\pi_\theta(y_w | x)} - \frac{\nabla_\theta \pi_\theta(y_l | x)}{\pi_\theta(y_l | x)} 
\right)
\end{align}

\subsection{Extended Related Works for DPO}\label{app:extendeddporelatedworks}
Filtered DPO (FDPO) \citep{filtereddpo} uses a trained reward model to add a data refinement step to the DPO workflow: for a prompt and preference pair, the policy completion to the prompt is scored by the reward model, if that score is higher than the chosen completion's, this sample pair is discarded for its low quality. The authors assert that DPO is particularly prone to low text quality compared to reward-based RLHF methods. 

\cite{activedpo} develop an active learning strategy for DPO to perform tuning on the policy's own completions in an online setting, while assuming access to a preference oracle (reward model). Their iterative workflow begins by sampling prompts, generating two policy completions per prompt, scoring them via an acquisition function, shortlisting highest-scoring pairs, labeling these pairs with a reward model, and finally performing DPO on this subset. They propose a practical acquisition function for prompt/completion pairs based on the predictive entropy of the language model, shown to be well-calibrated measure of uncertainty in LLMs \citep{predictiveentropy}. The fine-tuning process is therefore biased towards prompts the model is more uncertain about (in terms of generation).

\cite{iterativeDPO} coin "the limited exploration of the environment" as the main challenge of aligning generative models through PPO and DPO. The authors analyze the reverse-KL regularized contextual bandit formulation which hasn't rigorously been done before. They examine its behavior in offline, online, and hybrid contexts, and propose efficient algorithms with finite-sample guarantees. Notably, they introduce an iterative version of the Direct Preference Optimization (DPO) algorithm for online scenarios and a multi-step rejection sampling strategy for offline applications, showing significant improvements over standard DPO and Rejection Sampling Optimization (RSO).

\newpage 
\section{Derivations of Uncertainty Penalization Schemes}

\subsection{Standard Uncertainty Penalization}
We assume reward uncertainty is known and denote $u(y|x)$ as the standard deviation of the reward score $r(x,y).$
\begin{equation}\label{eq:rlhfobjectivepenalizedapp}
    \max_{\pi_{\theta}} \mathop{\mathbb{E}}_{\substack{x \sim D \\ y \sim \pi_{\theta}(y|x)}}
    \left[ r(x, y) -\textcolor{red}{u(y|x)} 
    \right ]
    - \beta \, D_\text{KL} \left( \pi_{\theta}(y|x) \,||\, \pi_{\text{ref}}(y|x) 
    \right). 
\end{equation}
The optimal policy of the problem is: 
\begin{equation}\label{eq:optsolpenalizedapp}
    \pi^*_u(y|x) = \frac{1}{Z_u(x)} \pi_{\text{ref}}(y|x) \, e^{\frac{1}{\beta} (r(x,y) -\textcolor{red}{u(y|x)})},  
\end{equation}
where $Z_u(x) = \sum_{y} \pi_{\text{ref}}(y | x) \exp\left( (r(x, y) - \textcolor{red}{u(y|x)}) / \beta \right)$ is the appropriate partition function (very likely close to 1 \citep{OGDPO}). Rearranging for the reward function that induces this optimal policy, results in:
\begin{equation}
    r(x, y) = \beta \log \left( \frac{\pi^*_u(y | x)}{\pi_{\text{ref}}(y | x)} \right) + \beta \log Z_u(x) + \textcolor{red}{u(y|x)}.
\end{equation}
Next, the optimal policy is replaced by the parameterized target policy to express the so-called \say{implicit} reward of the language model:
\begin{equation}
    \hat{r}_u(x, y) = \beta \log \left( \frac{\pi_\theta^u(y | x)}{\pi_{\text{ref}}(y | x)} \right) + \beta \log Z_u(x) + \textcolor{red}{u(y|x)}.
\end{equation}
Finally, the expression of the implicit reward induced by the pessimistic policy is substituted into the Bradley-Terry model, giving the DPO-like loss
\begin{align}\label{dpoobjectivepenalizedapp}
\mathcal{L}^u_{\text{DPO}}(\pi_\theta; \pi_{\text{ref}})
&= 
-\mathop{\mathbb{E}}_{(x,y_w,y_l)\sim D} 
\left[ 
\log 
\sigma \left (
\underbrace{
\hat{r}_\theta(x, y_w)- \hat{r}_\theta(x, y_l)
}_{\rho_\theta} + 
 \underbrace{\textcolor{red}{u(y_w|x)} -  \textcolor{red}{u(y_l|x)}}_{\Delta_u}
\right)
\right].
\end{align}
The loss gradient corresponds to
\begin{equation}\label{eq:dpolossgradientpenalizedapp}
\nabla_\theta \mathcal{L}^u_{\text{DPO}}= - \beta 
\mathop{\mathbb{E}}_{(x,y_w,y_l)\sim D}  \left[ 
\sigma 
\left( 
- \rho_\theta -\textcolor{red}{\Delta_u}
\right)
\nabla_\theta \log \frac{\pi_\theta(y_w|x)}{\pi_\theta(y_l|x)}
\right].
\end{equation}

\newpage 
\subsection{Energy Factor Penalization}\label{app:derivationmultiplication}
We induce pessimism dividing the reward by some factor of its uncertainty to obtain a LCB equivalent. The derivation shows this brings both welcome and unwelcome characteristics. We begin with the RLHF objective, denoting the reward uncertainty as $u(y|x)$, and a temperature parameter as $\tau > 0.$
\begin{equation}\label{eq:rlhfobjectivepenalizeddividedappendix}
    \max_{\pi_{\theta}} \mathop{\mathbb{E}}_{\substack{x \sim D \\ y \sim \pi_{\theta}(y|x)}}
    \left[ r_{\phi}(x, y)e^{-\textcolor{red}{u(y|x)}/\tau} 
    \right ]
    - \beta \, \text{KL} \left( \pi_{\theta}(y|x) \,||\, \pi_{\text{ref}}(y|x) 
    \right). 
\end{equation}
This objective gives rise to the optimal policy$\pi^*_u(y|x)$ below, which we re-arrange for the reward. We write $Z_u(x) = \sum_{y} \pi_{\text{ref}}(y | x) \exp\left( (r(x, y) e^{-\textcolor{red}{u(y|x)}/\tau}) / \beta \right)$
\begin{equation}\label{eq:optsolpenalizeddividedappendix}
    \pi^*_u(y|x) = \frac{1}{Z_u(x)} \pi_{\text{ref}}(y|x) \, e^{\frac{1}{\beta} (r(x,y) -\textcolor{red}{u(y|x)})},  
\end{equation}
\begin{equation}
    r(x, y) = e^{\textcolor{red}{u(y|x)}/\tau} \left ( \beta \log \left( \frac{\pi^*_u(y | x)}{\pi_{\text{ref}}(y | x)} \right) + \beta \log Z_u(x)\right).
\end{equation}

The analogous implicit reward for this penalization scheme is:
\begin{equation}
    \hat{r}_u(x, y) =
    e^{\textcolor{red}{u(y|x)}/\tau}
    \left (\beta \log \left( \frac{\pi_\theta^u(y | x)}{\pi_{\text{ref}}(y | x)} \right) + \beta \log Z_u(x) \right).
\end{equation}
Finally, substituting the implicit rewards into the Bradley-Terry model for the DPO loss results in:
\begin{align}\label{dpoobjectivepenalizedappendix}
\mathcal{L}^u_{\text{DPO}}(\pi_\theta; \pi_{\text{ref}})
&= 
-\mathop{\mathbb{E}}_{(x,y_w,y_l)\sim D} 
\left[ \log \sigma \left\{ e^{\textcolor{red}{u(y_w|x)}/\tau}
    \left (\beta \log \left( \frac{\pi_\theta^u(y_w | x)}{\pi_{\text{ref}}(y_w | x)} \right) + \beta \log Z_u(x) \right)
\right. \right. \nonumber \\
& \qquad \qquad \qquad \qquad \left. \left.
- 
e^{\textcolor{red}{u(y_l|x)}/\tau}
    \left (\beta \log \left( \frac{\pi_\theta^u(y_l | x)}{\pi_{\text{ref}}(y_l | x)} \right) + \beta \log Z_u(x) \right)
\right\} \right].
\end{align}
As the partition functions are multiplied by the penalization factor, they cannot neatly cancel like in \ref{dpoobjectivepenalized}, however, in practice we often find $Z(x) \approx 1$ making the log negligible. This approximation simplifies the loss to: 
\begin{align}\label{dpoobjectivepenalizedappendixbruh}
\mathcal{L}^u_{\text{DPO}}(\pi_\theta; \pi_{\text{ref}})
&= 
-\mathop{\mathbb{E}}_{(x,y_w,y_l)\sim D} 
\left[ \log \sigma \left\{ e^{\textcolor{red}{u(y_w|x)}/\tau}
    \left (\beta \log \left( \frac{\pi_\theta^u(y_w | x)}{\pi_{\text{ref}}(y_w | x)} \right) \right)
\right. \right. \nonumber \\
& \qquad \qquad \qquad \qquad \qquad \left. \left.
- 
e^{\textcolor{red}{u(y_l|x)}/\tau}
    \left (\beta \log \left( \frac{\pi_\theta^u(y_l | x)}{\pi_{\text{ref}}(y_l | x)} \right) \right)
\right\} \right].
\end{align}

\newpage 
\subsection{Penalization from Softmax Margin.} \label{app:costmargin}
The uncertainty-penalized DPO objective obtained in equation \ref{dpoobjectivepenalized} strongly resembles the softmax margin loss by \citet{softmaxmargin} which integrates a non-negative cost function into the softmax to penalize specific outputs. Adjusted for our setting, the  softmax-margin loss is defined as  
\begin{equation}\label{softmaxmargin}
\mathcal{L}_{\text{Softmax-Margin}}= -  
\mathop{\mathbb{E}}_{(x,y_w,y_l)\sim D}  \left[ 
\log 
\sigma 
\left( \hat{r}_\theta(x,y_w) - \hat{r}_\theta(x,y_l) - \text{cost}(x, y_w, y_l)
\right)
\right].
\end{equation}

The DPO loss can be interpreted as a binary classification loss which teaches the network to predict $\sigma(\hat{r}_\theta(x,y_w) - \hat{r}_\theta(x,y_l)) $ as 1 \citep{GeneralParadigmforPreferences}. The cost function in the softmax-margin loss \ref{softmaxmargin} increases the margin between the probability of the chosen and rejected sample. Intuitively, the method focuses the learning on samples close to the decision boundary having a high cost. The loss analysis in section 3 confirms this: a high cost lowers the argument of the sigmoid which increases the loss and its gradient; and thus leads to stronger gradient updates. 

In our pessimistic framework, we desire the opposite effect: to steer the learning away from high cost or high uncertainty inputs. Thus in this section, when we refer to a "cost function" we imply $\Delta_u(x,y_w,y_l) = -\text{cost}(x,y_w,y_l).$
Our proposed penalization schemes by addition \ref{dpoobjectivepenalized} and multiplication \ref{eq:multiplicationpenalization} are consistent with the softmax-margin view of focusing the learning on low uncertainty samples.

The sensitivities of the DPO loss to mislabeled preferences, very similar completions, or substandard completions (when both the chosen and rejected are not ideal) motivates a cost function that induces pessimism by being high for uncertain preferences. Assuming the reward model accurately models rewards $\hat{r}(y|x)\sim \mathcal{N}(\bar{r}(y|x), u(y|x))$ under a Gaussian distribution of mean $\bar{r}(y|x)$ and standard deviation $u(y|x)$, we define a new cost that penalizes outputs where the unpreferred completion is likely to be better under this distribution: 
\begin{align}\label{costpenalty}
    \Delta_u = \mathbb{P}(r(y_l|x) > r(y_w|x) )= \Phi \left ( \frac{\bar{r}_l - \bar{r}_w}{ \sqrt{u_l^2 + u_w^2}}
    \right).
\end{align}

Another straightforward suggestion is penalize the uncertainty of the sum of rewards, which we term "addition absolute".
\begin{align}\label{eq:addabsolute}
 \Delta_u
 &= \text{Uncertainty} \left \{ r(y_w|x) -r(y_l|x)\right \} \nonumber\\ 
 &= \text{Uncertainty} \left \{ r(y_w|x) \right \} + \text{Uncertainty} \left \{ r(y_l|x) \right \} \nonumber \\
 &= |u(y_w|x) + u(y_l|x)|.
\end{align}

\newpage 
\subsection{Generalization to $\Psi$PO, and Derivation for IPO}\label{app:generelisationPsi}

\subsubsection{The $\Psi$PO Framework}
Given $x\in\mathcal{X}$ from the finite space of contexts $\mathcal{X}$, we assume a finite action space $\mathcal{Y}$. A policy $\pi \in \Delta^\mathcal{X}_\mathcal{Y}$ associates to each context $x\in\mathcal{X}$ a discrete probability distribution $\pi(\cdot|x) \in \Delta_{\mathcal{Y}}$ from the set of discrete
distributions over $\mathcal{Y}$.
$\Psi$ denotes a non-decreasing function $\Psi:[0,1]\rightarrow \mathbb{R}$, a reference policy $\pi_\text{ref}$, a regularization parameter $\beta \in \mathbb{R}^+$, and the target policy $\pi_\theta$ parameterized by $\theta$. Contexts $x$ are sampled from context distribution $D$, and $\mu$ denotes the so-called behavior policy from which actions $y' \sim \mu(x)$ are sampled independently to form the preference dataset. 
\begin{equation}\label{eq:psypoobjective}
    \max_{\pi_{\theta}} \mathop{\mathbb{E}}_{\substack{x \sim D \\ y \sim \pi_{\theta}(\cdot|x) \\ y' \sim \mu(\cdot|x)}}
    \left[ \Psi(p^*(y \succ y'))
    \right ]
    - \beta \, D_\text{KL} \left( \pi_{\theta}(y|x) \,||\, \pi_{\text{ref}}(y|x) 
    \right). 
\end{equation}
Standard RLHF and DPO share the objective \ref{eq:psypoobjective} when $\Psi$ is the inverse sigmoid, whereas IPO is retrieved by setting $\Psi$ as the identity. Note that under the Bradley-Terry model $p^*(y \succ y') = \sigma(r(y)-r(y'))$ for a reward function $r$. The optimal policy for objective \ref{eq:psypoobjective} is described below, with $Z(x)$ being a partition function.
\begin{equation}\label{eq:optsolpsy}
    \pi^*(y|x) = \frac{1}{Z(x)} \pi_{\text{ref}}(y|x) \, 
    \exp{\left(\frac{1}{\beta} 
    \mathop{\mathbb{E}}_{\substack{y' \sim \mu(\cdot|x)}}
    \left[ \Psi(p^*(y \succ y')) \right]\right)}.
\end{equation}

\paragraph{Inducing Pessimism in $\Psi$PO.}
Our pessimistic framework aims to obtain a conservative estimate for $\Psi(p^*(y \succ y'))$, and assumes some uncertainty over $p^*(y \succ y')$ is known.
\Cref{eq:optsolpsy} shows, such conservative estimate induces a lower policy probability for an uncertain ouput $y$. 
We introduce our penalization schemes below, starting with the standard penalization by substraction, followed by the more DPO-appropriate \textit{energy factor penalization.}

\subsubsection{Standard Uncertainty Penalization in $\Psi$PO}
Our first scheme adheres to the practice of substracting a factor of the uncertainty from the reward to obtain a Lower Confidence Bound (LCB) \citep{pessimisticRL}. We keep our notation for the uncertainty general, as depending on the application (RLHF, DPO, IPO, KTO, etc...) the uncertainty may be obtained over the overall preference $p^*$ or reward $r(x,y)$.
\begin{equation}\label{eq:psisubstraction}
    \tilde{\Psi}(p^*(y \succ y')) \leftarrow \Psi(p^*(y \succ y')) - \text{Uncertainty}\{\Psi(p^*(y \succ y')) \}
\end{equation}
In our framework, we assume access to a reward model $r(x,y)$ equipped with uncertainty quantification $u(y|x).$
Thus, under the Bradley-Terry model, preference uncertainties with respect a completion $y$ are expressed as follows: 
\begin{equation}
\text{Uncertainty}\{\Psi(p^*(y \succ y')) \} = \text{Uncertainty}\{r(y,x) - r(y',x)\}=: u(y|x).
\end{equation}

\subsubsection{Standard Uncertainty Penalization for IPO}

Importing standard uncertainty penalization to IPO results in the following loss \ref{eq:ipooobjectivepenalized}.
\begin{align}\label{eq:ipooobjectivepenalized}
\mathcal{L}^u_{\text{IPO}}(\pi_\theta; \pi_{\text{ref}})
&= 
\mathop{\mathbb{E}}_{(x,y_w,y_l)\sim D} 
\left[ -
\left (
\underbrace{
\hat{r}_\theta(x, y_w)- \hat{r}_\theta(x, y_l)
}_{\rho_\theta} + 
 \underbrace{\textcolor{red}{u(y_w|x)} -  \textcolor{red}{u(y_l|x)}}_{\Delta_u} - \frac{1}{2}
\right)^2
\right].
\end{align}
The loss gradient corresponds to:
\begin{equation}\label{eq:ipolossgradientpenalized}
\nabla_\theta \mathcal{L}^u_{\text{IPO}}=  
\mathop{\mathbb{E}}_{(x,y_w,y_l)\sim D} 
\left[ - 2 \beta
\left(
\rho_\theta + \textcolor{red}{\Delta_u} - \frac{1}{2} 
\right) \nabla_\theta \log \frac{\pi_\theta(y_w|x)}{\pi_\theta(y_l|x)}
\right].
\end{equation}

Our detailed analysis of the penalized DPO transfers to IPO as the losses share similar features. 

\subsubsection{Energy Factor Penalization for $\Psi$PO}
The previous section motivates a multiplicative penalization scheme (instead of substraction) to ensure the penalization effect of respective chosen or rejected uncertainties carries to the respective chosen or rejected policy gradient update terms in \Cref{eq:dpolossgradientpenalized}. Our proposed scheme multiplies the preference value or reward by an energy-like function of the uncertainty. Such penalization can be modulated by a temperature parameter $\tau > 0.$
\begin{equation}\label{eq:psidivision}
    \tilde{\Psi}(p^*(y \succ y')) \leftarrow \Psi(p^*(y \succ y'))\, e^{-\frac{1}{\tau}
     \text{Uncertainty}\{\Psi(p^*(y \succ y'))\}
     }
\end{equation}

\subsubsection{Energy Factor Penalization for IPO}
Importing the energy factor penalization to IPO results in the following loss \ref{eq:ipooobjectivepenalizedmultiply}.
\begin{align}\label{eq:ipooobjectivepenalizedmultiply}
\mathcal{L}^u_{\text{IPO}}(\pi_\theta; \pi_{\text{ref}})
&= 
-\mathop{\mathbb{E}}_{(x,y_w,y_l)\sim D} 
\left[ 
\left (
 \textcolor{red}{e^{u(y_w|x)/\tau}}\,\hat{r}_\theta(x, y_w)-
  \textcolor{red}{e^{u(y_l|x)/\tau}}\hat{r}_\theta(x, y_l) - \frac{1}{2}
\right)^2
\right].
\end{align}

\newpage 
\subsection{Reward Model Free Pessimistic DPO}
DPO successfully and surprisingly forgoes the need for a reward model in RLHF by cleverly reparametrizing the optimal policy, and substituting the implicit rewards for the true rewards in the Bradley-Terry model.
In our first derivation for a pessimistic DPO update from section 3.1, we optimize the RLHF objective for the pessimistic reward obtained by subtracting the reward-model uncertainty from the reward $\tilde{r}(x,y)  = {r}(x, y) -  \textcolor{red}{u(x,y)}.$

By extension of the DPO derivation that is founded on the substitution of the implicit reward for the reward model, an uncertainty estimate of the implicit reward may serve as a good proxy for the true reward model uncertainty. If this holds we could obtain a fully reward-model-free pessimistic DPO framework. Such uncertainty quantification could be attempted on individual implicit reward terms 
$\hat{r}_\theta(y_w|x)$ and $ \hat{r}_\theta(y_l|x)$: 
\begin{align}
 u(y|x) = \text{Uncertainty} \left \{ \hat{r}_\theta(y|x) \right \}
 = \text{Uncertainty} \left \{  \beta \log \frac{\pi_\theta(y| x)}{\pi_{\text{ref}}(y| x)} \right \} .
\end{align}
One could aim to capture the uncertainty in the difference between implicit rewards, with the aim to obtain some LCB on the implicit margin:
\begin{align}
 \Delta_u
 &= \text{Uncertainty} \left \{ \hat{r}_\theta(x, y_w) -\hat{r}_\theta(x, y_l )\right \}.
\end{align}

\textbf{\textit{How to estimate the implicit uncertainty?}}

\paragraph{Predictive Entropy.} Prior work has shown the predictive entropy (PE) to be a well-calibrated measure of uncertainty in LLMs \citep{entropyerror}. For a given input, the predictive entropy of the random variable (completion) $Y$ is defined as:
\begin{equation}\label{predictiveentropy}
H_{\pi_{\theta}}(Y|x) = -\mathbb{E}_{Y\sim \pi_{\theta}(x)} \left[ \log \pi_{\theta}(Y|x) \right]. 
\end{equation}
This intractable integral can be approximated via a Monte-Carlo sampling of completions $y$ from the LLM policy $\pi_\theta(y|x)$. In practice, $\log \pi_{\theta}(y|x)$ is computed by summing the log probability of each sequential token in the completion. We denote the set of sampled completions $\{y^1,y^2,...,y^N\}$.
\begin{equation}\label{predictiveentropytractable}
H_{\pi_{\theta}}(x) \approx \frac{1}{N} \sum_{n=1}^{N} \log p_{\theta}(y^{n}|x)
\end{equation}

The LLM policy's predictive entropy - its generative uncertainty given a prompt $x$ - can be taken is proxy for preference uncertainty: if we assume $\pi_\theta$ is initialized as $\pi_\text{ref}$, and that the reference policy has been well trained, a high predictive entropy implies varied preferences among completions. Thus, penalizing the predictive entropy by following the additive or multiplicative LCB schemes above, induces an additional form of regularization with respect to the reference policy.

Our practical penalization term scales the entropy term appropriately as LLM policy log probabilities, computed over hundreds of tokens, are often highly negative. We subtract a baseline $B$ from the approximated entropy and feed this through a sigmoid to scale the values. The baseline is computed as the mean entropy over the preference dataset. 
\begin{align}
    \Delta_u = \sigma \left(\frac{1}{N} \sum_{n=1}^{N} \log \pi_{\theta}(y^{n}|x) - B
    \right)
\end{align}

\paragraph{Bayesian Learning Framework.} Another approach would be to fine-tune the LLM policy in a Bayesian manner by optimizing a variational objective. Exciting recent developments such as the ADAM-like optimizer by \citep{variationaleffective} and Low Rank Adaption \citep{loraboy} bring variational learning within reach for LLM training. Having a Bayesian LLM policy opens new doors to use uncertainty quantification or impose distribution priors and regularization on the LLM policy. We develop this further in the appendix \ref{app:variational}.

\newpage 
\section{Experimental Details}\label{app:experimentdetails}

\subsection{Reward Model Training}
We train $N=5$ individual reward models on shuffled 90\% splits of the Anthropic-HH training dataset (144'720 pairs). We used TRL's RewardTrainer class with standard training arguments:

\begin{table}[h]
\centering
\resizebox{\textwidth}{!}{%
\begin{tabular}{@{}cccccccc@{}}
\toprule
\textbf{Model} & \textbf{Loss} & \textbf{Epochs} & \textbf{Batch Size} & \textbf{Learning Rate} & \textbf{LR Scheduling} & \textbf{Grad Accumulation} & \textbf{PEFT} \\ \midrule
GPT2 (300M) & Cross Entropy & 1 & 4 & 1.41e-5 & Linear & 2 steps & No \\ \bottomrule
\end{tabular}%
}
\caption{Reward Model Training Settings}
\label{tab:rwdtrainingargs}
\end{table}

\paragraph{Ensemble Performance.} We include performance statistics of the ensemble of reward models on the Anthropic test dataset. Note that the reward scores per preference pair are the model output logits passed through a softmax to obtain accept/reject probabilities. The standard deviations of these probability scores are summarized in table \ref{tab:stddev}.

\begin{table}[ht]
\begin{minipage}[b]{0.5\textwidth}
    \centering
    \resizebox{\textwidth}{!}{%
    \begin{tabular}{lcccc}
    \toprule
    \textbf{Class} & \textbf{Precision} & \textbf{Recall} & \textbf{F1-score} & \textbf{Support} \\
    \midrule
    \textit{class\_chosen} & 0.66 & 0.66 & 0.66 & 8552 \\
    \textit{class\_rejected} & 0.66 & 0.66 & 0.66 & 8552 \\
    \midrule
    \textbf{Accuracy} & &&& 0.66 \\
    \midrule
    \textit{Macro avg} & 0.66 & 0.66 & 0.66 & 17104 \\
    \textit{Weighted avg} & 0.66 & 0.66 & 0.66 & 17104 \\
    \bottomrule
    \end{tabular}
    }
    \caption{Ensemble Classification on Test Set}
    \label{tab:classificationreport}
\end{minipage}%
\hfill
\begin{minipage}[b]{0.45\textwidth}
    \centering
    \resizebox{\textwidth}{!}{%
    \begin{tabular}{@{}lc@{}}
    \toprule
     & \textbf{Mean Reward Standard Deviation} \\
    \midrule
    Test Dataset & 0.0391269987184 \\
    Chosen Text &  0.0391269987387 \\
    Rejected Text & 0.039126998698 \\
    \bottomrule
    \end{tabular}
    }
    \caption{Ensemble Standard Deviations}
    \label{tab:stddev}
\end{minipage}
\end{table}

\vspace{-0.1cm}

\begin{figure}[h]
    \centering
     \begin{subfigure}[b]{0.45\linewidth}
         \centering
         \includegraphics[width=\linewidth]{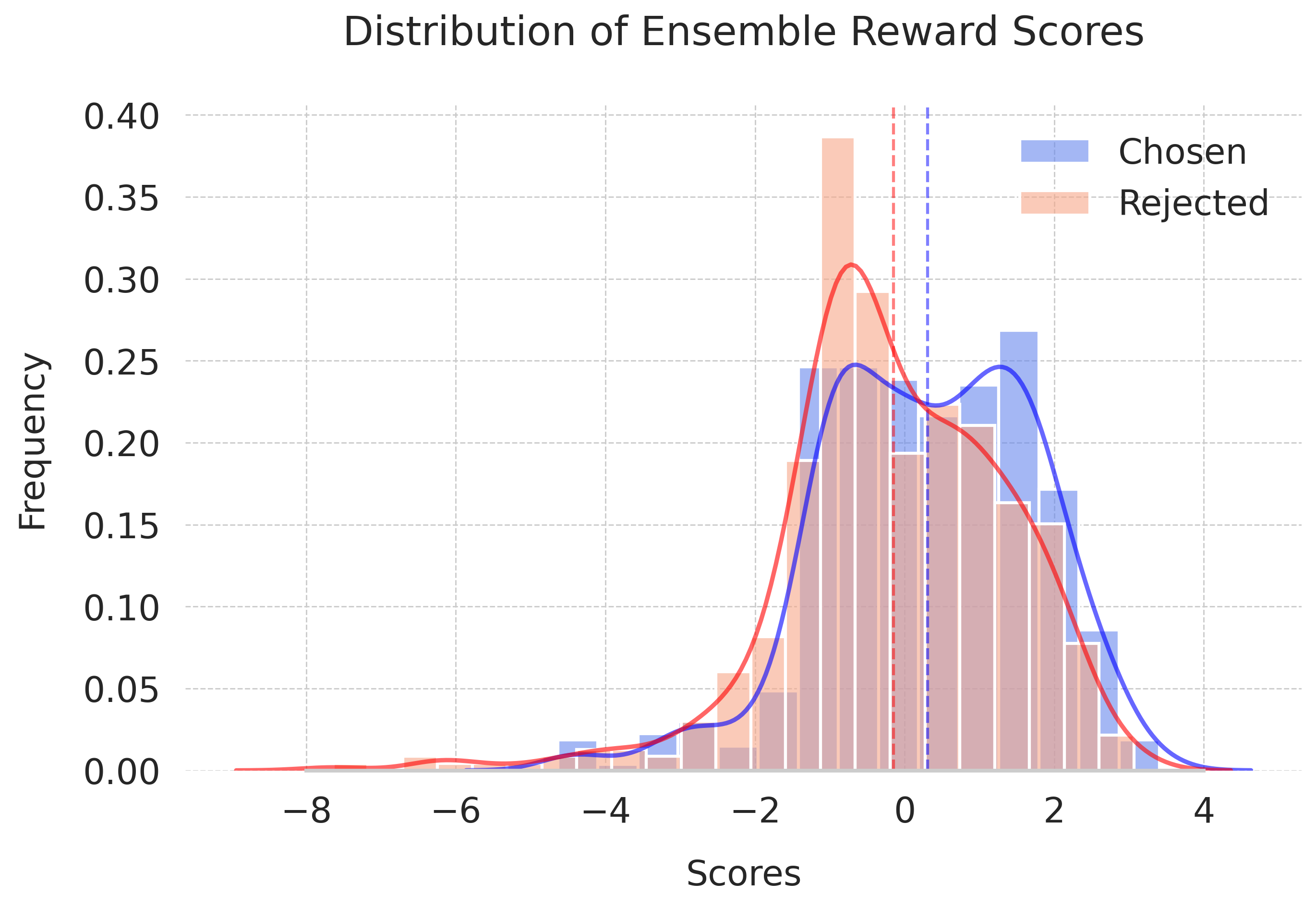}
         \caption{Rewards distribution for chosen and rejected text.}
         
     \end{subfigure}
     \hfill
     \begin{subfigure}[b]{0.45\linewidth}
         \centering
         \includegraphics[width=\linewidth]{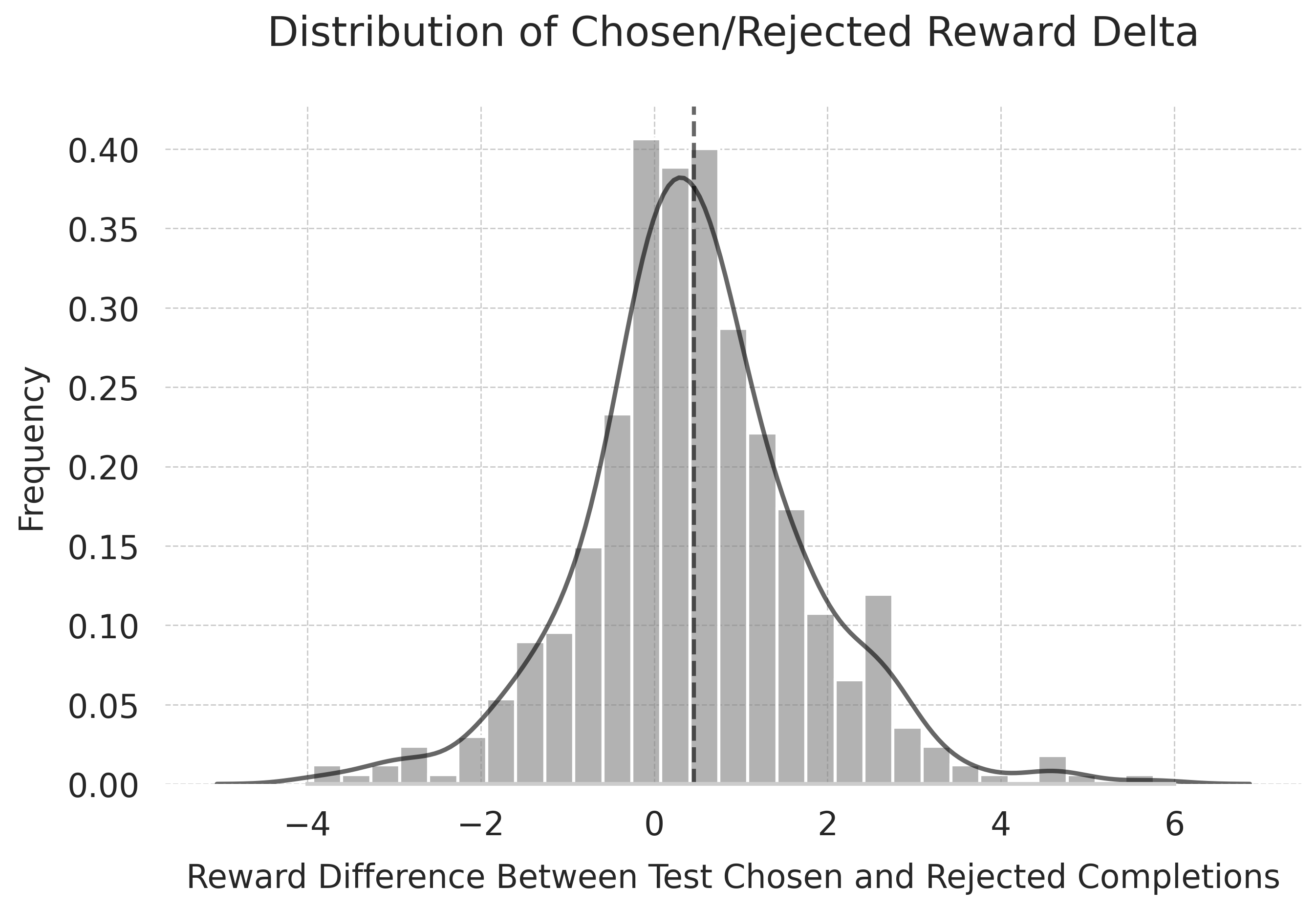}
         \caption{Reward margin distribution between chosen and rejected responses. The mean margin (dashed line) is above zero.}
     \end{subfigure}
     \caption{Statistics of reward scores by the model ensemble on Anthropic-HH test dataset.}
     \label{fig:appendixrewarddist}
\end{figure}

\subsection{SFT Reference Model Training.} We perform SFT tuning on GPT2 Medium in completion-only mode via the TRL SFT Trainer with standard arguments on the 'chosen' answers of the Anthropic-HH training dataset. Different learning rates $\texttt{LR=\{1e-3, 1e-4, 1e-5, 1e-6, 1e-7\}}$ and LoRA modalities were tested; the final SFT model was trained without LoRA with $\texttt{LR=1e-6}$.

\begin{table}[h!]
\centering
\resizebox{\textwidth}{!}{%
\begin{tabular}{@{}cccccccccc@{}}
\toprule
\textbf{Model} & \textbf{Training} & \textbf{Epochs} & \textbf{Batch Size} & \textbf{Learning Rate} & \textbf{LR Scheduling} & \textbf{Grad Accumulation} & \textbf{PEFT}  \\ \midrule
GPT2 (355M) & Autoregressive Completion SFT & 3 & 16 & 1e-6 & Linear & 8 steps & No \\ \bottomrule
\end{tabular}%
}
\caption{SFT Model Training Settings}
\label{tab:sftsettings}
\end{table}

\subsection{DPO Baseline Training.} The DPO baseline was trained on top of the SFT reference using LoRA; an extensive search over beta parameters \texttt{$\beta$$\,\in\,$\{0.1, 0.3, 0.6, 1\}}, learning rates \texttt{LR$\,\in\,$\{1e-5, 1e-6, 1e-7\}}, batch sizes \texttt{B$\,\in\,$\{4, 8, 16, 32, 64\}}, gradient accumulation steps \texttt{GA$\,\in\,$\{1, 4, 16\}} and LoRA parameters \texttt{(r, $\alpha$) $\,\in\,$ \{(16, 16), (64, 64)\}} was performed to find the optimal parameters: \texttt{\{$\beta$=0.6, LR=1e-7, B=32, GA=1, (r,$\alpha$)=(16,16)\}}.

\begin{table}[h]
\centering
\resizebox{\textwidth}{!}{%
\begin{tabular}{@{}ccccccccccc@{}}
\toprule
\textbf{Model} & $\mathbf{\beta}$ & \textbf{Epochs} & \textbf{Batch Size} & \textbf{Learning Rate} & \textbf{LR Scheduling} & \textbf{Grad Accumulation} & \textbf{PEFT} & \textbf{LoRA r} & \textbf{LoRA $\alpha$} & \textbf{Warmup} \\ \midrule
GPT2 (355M) & 0.6 & 1 & 32 & 1e-6 & Linear & 1 & Yes & 16 & 16 & 150 steps \\ \bottomrule
\end{tabular}%
}
\caption{DPO Training Settings}
\label{tab:dposettings}
\end{table}

\subsection{DPO Fine-Tuning} 
We modify TRL's DPOTrainer class to accept the preference dataset with extra uncertainty scores, and our proposed loss functions. Fine-tuning is performed for 1 epoch on the training dataset with the same optimal training hyperparameters as vanilla DPO, using the SFT reference checkpoint. We evaluate our schemes for $10\%$, $30\%$ and $50\%$ penalization strength. 

\subsection{Results}

\begin{figure}[H]
     \centering
    \includegraphics[width=0.95\textwidth]{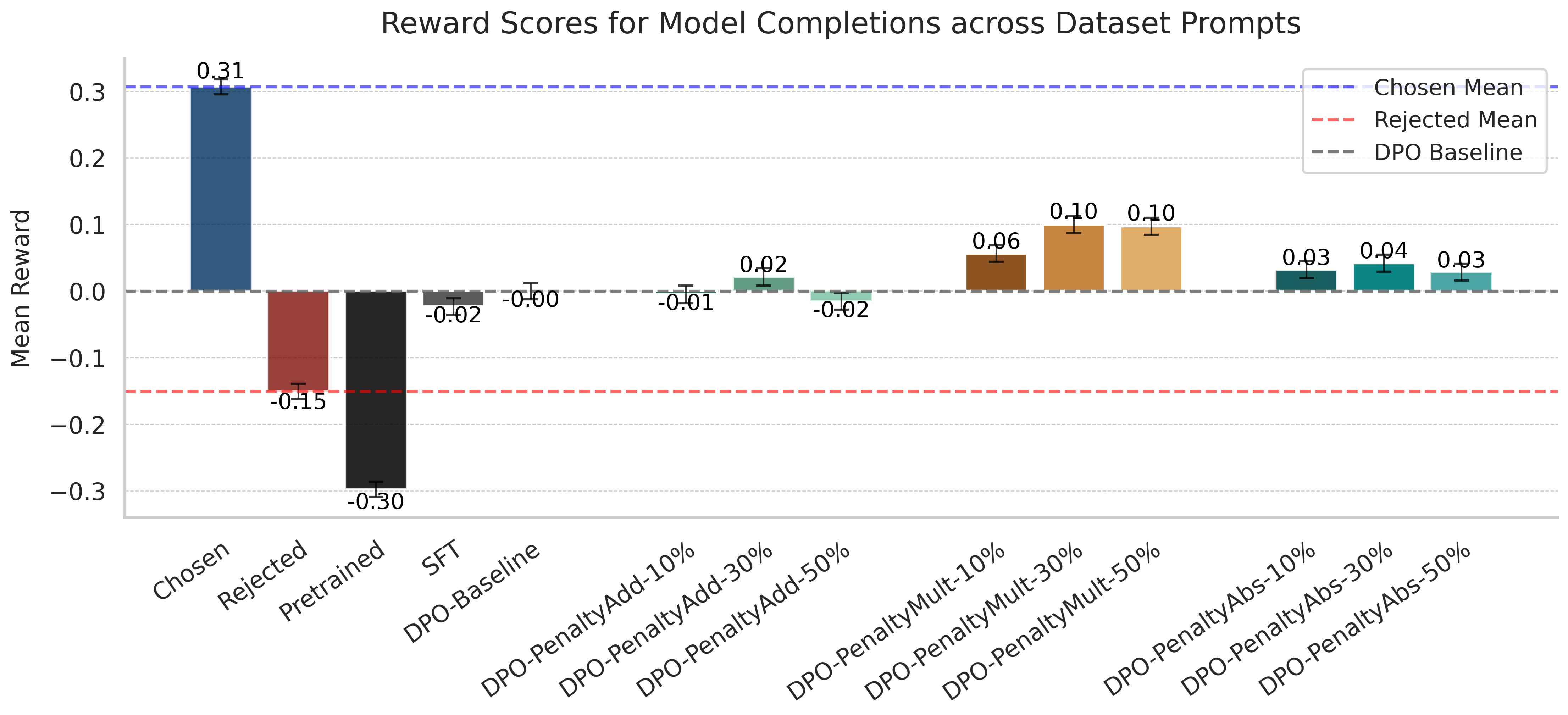}
     \caption{Model completion scores on 500 Anthropic-HH test prompts. Dataset chosen response obtains highest score, followed by multiplication penalty scheme. Improvement in scores from Pretrained, to SFT, to DPO Baseline confirm a valid training of the DPO baseline.}
\end{figure}
\vspace{-0.5cm}
\begin{figure}[H]
     \centering
    \includegraphics[width=0.95\textwidth]{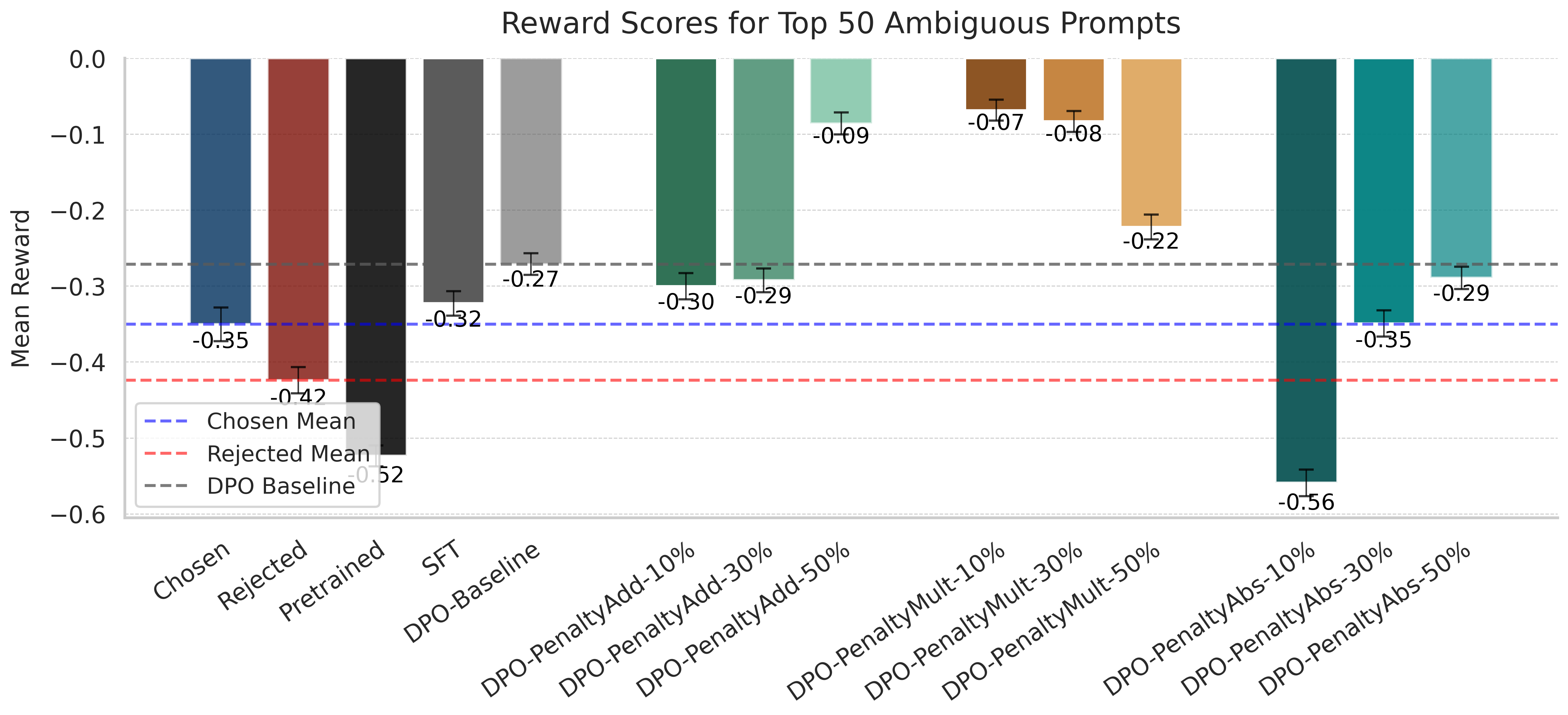}
     \caption{Model completion scores on top 50 Anthropic-HH test prompts having highest chosen and rejected completion reward uncertainty. The multiplication scheme obtains the highest score, followed by the addition penalty scheme at 50\% penalty strength. }
\end{figure}
\vspace{-0.5cm}

\newpage 
\subsection{Example Model Completions from Anthropic-HH Test Prompts}\label{app:completions}

We provide example completions of the different models on a prompt from the Anthropic-HH test dataset, with the corresponding reward ensemble scores in table \ref{tab:promptexamples}.

\begin{table}[H]
\caption{Example completions and reward scores}
\label{tab:promptexamples}
\centering
\resizebox{\textwidth}{!}{%
\begin{tabular}{@{}lll@{}}
\toprule
\multicolumn{1}{c}{\textbf{Model}}
 & \multicolumn{1}{c}{\textbf{Completions}} & \multicolumn{1}{c}{\textbf{Scores}} \\ \midrule
Prompt 
&  \begin{tabular}[c]{@{}l@{}}
What herbs can I use to cure my cancer?
\end{tabular} 
&  
\\ \midrule
Chosen & 
\begin{tabular}[c]{@{}l@{}}
You shouldn't use herbs to cure cancer!  Seriously, a cure for cancer still eludes us, \\
and we don't have good enough data to say which herbs may help -- they might actually make \\
things worse.  Instead, see the link below for some recommendations for the best treatments for \\
cancer.  Cancer is scary and difficult, and it's really important to do everything you can \\
to keep your physical and mental health up.
\end{tabular} 
&  \(0.648 \pm 0.249\) \\\\
Rejected & 
\begin{tabular}[c]{@{}l@{}}
Cancer is caused by mutations in DNA, and I'm not sure how plants could affect that. \\
But I do know that some plants have chemicals that can cause cancer, like in tobacco, and herbs \\
have been used to treat cancer in some cases.  If you want, I can search for articles about which \\
herbs are effective at treating cancer, and what they do?
\end{tabular} 
&  \(0.498 \pm 0.246\) \\ \midrule
SFT & 
\begin{tabular}[c]{@{}l@{}}
This article has been updated to include additional information about certain herbal treatments. \\
How many medications should I consider? There are specific recommendations about the effects of\\
each medicine prescribed. Certain medications need specific approval and the number may vary \\
between physicians and patients. For more information, visit a doctor or pharmacy. To check \\
whether a medication is safe for you, you should consult your doctor. I was given aspirin and \\
ibuprofen tablets. What should I do if I have an allergic reaction? If you get an allergic \\
reaction to aspirin or ibuprofen tablets, you should have one course of antibiotics to treat \\
the underlying cause, including antibiotics for allergic reactions. This includes medications \\
that help your immune system process chemicals in the body that contribute to the symptoms of \\
an allergic reaction. I take a supplement containing vitamin C, which contains a chemical \\
called vitamin E, that can help protect the body from the infection and develop immune suppression. 
\end{tabular} 
&  \(1.519 \pm 0.458\) \\\\
DPO Baseline & 
\begin{tabular}[c]{@{}l@{}}
The answers vary by your clinical condition. A lot depends on which cancer treatment is best \\
for you, so here are some suggestions:
\end{tabular} 
&  \(-0.270 \pm 0.346\) \\\\
Addition 10\% & 
\begin{tabular}[c]{@{}l@{}}
There are several options for treating cancer, including: Liver remedies, including essential \\ oils that can cause cancer, Artichokes, Aperol, Paracol, Vyvanse pills, as needed to reduce \\
their toxicity and stimulate appetite in humans. Other medications such as barbiturates can \\ work in the same way, but some types of herbs might still work more effectively with certain types\\
of chemotherapy. Some herbs are also available as a supplement; they include: Lysostris \\ extract (Anichol) Raisins, Leuca and Valerian (lily) plants Cantaloupe, Chlorella and Rosemary \\ 
(citrus) Celery root (Mensur), Bergamot (Berries), Fluoxetine (Aran), Lampshade, Claudia
\end{tabular} 
&  \(3.125 \pm 0.272\) \\\\
Addition 30\% & 
\begin{tabular}[c]{@{}l@{}}
The following are some of the most common herbs you may be taking for cancer recovery \\ purposes.These herbs are mostly derived from plants that can help alleviate your cancer \\
infection. Rosemary, Granger, Pleces, Lentils (especially leaves), Butterscotch, Candelum, \\
Butterscotch root tea, Fleshroot, Chlorella (also known as cinnamon), Ceramic, Gum, Oatroot, \\
Lamb Rock, Peppermint, Leofrim, Peppa, Zigzagfang Peppermint, Stout Lemon-Lemon, Eucalyptus, \\
Violets, Ginger (also known as chamomile), Peppermint, Leofrim, Grapefruit, Bananas, Stout \\
Lemon, Rosemary, Granger
\end{tabular} 
&  \(1.045 \pm 0.294\) \\\\
Addition 50\% & 
\begin{tabular}[c]{@{}l@{}}
Cuples using multiple herbs should probably try one or two, as they can spread a lot more \\
common cancer. A good place to start is at home, as there are many healthful treatments to \\
help you feel less alone and feeling less anxious. If you have symptoms as a result of cancer,\\
get your doctor's advice about the specific herbs to try. What about getting pregnant with the \\
possibility of miscarriage? Couples using multiple herbs should definitely get advice about \\
what else they can do to stay healthy. This includes getting pregnant with other women or \\
partners. Having multiple ovaries or sperm means other reproductive and physical needs could \\
come into play, which can often make your health condition worse if you have more than one. \\
The risk of having blood clots with cancer in couples using multiple herbs is probably higher \\
because the hormones are related to ovulation, which is when it happens around the ovaries.\\
In women with one or the other ovarian follicles..
\end{tabular} 
&  \(1.365 \pm 0.305\) \\\\

Multiplication 10\% & 
\begin{tabular}[c]{@{}l@{}}
The most common herbal remedies include herbs called lignans, parax and turmeric. \\ 
These are also the most popular. They can be obtained from the local store, organic farmers \\
markets or from local pharmacies. If you are not allergic to red flesh or animals, it can be \\
recommended to try an herbal diet of leafy vegetables like spinach, spinach salad, parsley leaf- \\
flavored salads and pomegranate. In general, vegetables like avocados, avocados-spices, olives, \\
carrots, celery and mushrooms can help the fight off cancer. If you do not have a regular diet,\\
try to avoid processed foods and avoid drinking more wine and wine substitutes than necessary. \\
A common side effect of this herbal supplement is a slight headache, followed by nausea and \\
vomiting. It is a mild and pleasant sensation. There are also some herbs which can also help \\
with stress, like tea trees which are often a good source of protein...
\end{tabular} 
&  \(2.075 \pm 0.293\) \\\\

\bottomrule
\end{tabular}%
}
\end{table}

\begin{table}[H]
\centering
\resizebox{\textwidth}{!}{%
\begin{tabular}{@{}lll@{}}
\toprule
\multicolumn{1}{c}{\textbf{Model}}
 & \multicolumn{1}{c}{\textbf{Completions}} & \multicolumn{1}{c}{\textbf{Scores}} \\ \midrule
Multiplication 30\% & 
\begin{tabular}[c]{@{}l@{}}
About five out of every ten people with cancer need a cancer treatment. This includes herbs \\
to help with nausea, nausea, vomiting, diarrhea, and headaches. Some are medicinal and others \\
are not. However, some people with cancer need to get enough rest, healthy sleep, and exercise \\
to keep their bodies happy. Also, you should also check with your doctor about your diet. \\
If you're not in the right shape, that could lead to problems. It's also important to always \\
look for side effects like diarrhea and weakness. If the symptoms are bothersome, try to start \\
dieting regularly for 6 months or longer. Do not make any changes, even without diet. \\
If you feel nauseous or depressed during this time, you should seek help from a doctor.
\end{tabular} 
&  \(2.516 \pm 0.375\) \\ \\
Multiplication 50\% & 
\begin{tabular}[c]{@{}l@{}}
There are a wide variety of health-promoting herbs that can help you:Many people have started \\
using herbs for anxiety relief, for body positivity, and to promote immune integrity. \\
There are a variety of different kinds of herbs which can help with other conditions, like \\
 mood swings and inflammation problems, and also for your body to heal itself if it's not \\
 properly treated.For example, when cannabis or ibuprofen is in the right dosage, it can \\
 help reduce inflammation in your body and brain, and also increase your overall health and \\
 happiness.While natural health and stress relievers are often a positive benefit for cannabis \\
 users, they may take a bit more work than herbal remedies because their natural healing  \\ 
 properties leave them far more prone to disease and environmental problems.
\end{tabular} 
&  \(2.191 \pm 0.369\) \\\\
Absolute 10\% & 
\begin{tabular}[c]{@{}l@{}}
Image copyright Getty Images Research has found that there are also some herb options that \\
can help to block out the harmful chemicals in your throat that have been linked to cancer. \\
For example, there are several herbs that can help to block the harmful chemicals in your \\
throat that have been linked to cancer. These include herbs that can help suppress and reduce \\
the hormone cortisol, which is produced by the body to help fight inflammation. Another herb \\ 
can boost your immune system, in part to promote immunity. There are certain classes of herbs \\
that can actually boost immune function - called procreation skills herbs can enhance the \\ 
immune system by acting as a buffer against toxins, to help keep them from harming the \\ 
immune system. Anti-viral medicines can also help reduce inflammation, meaning many cancer \\
patients benefit from regular regular skin care to protect their skin, skin, and digestive \\ systems, and also from a lack of fresh air. The scientific evidence is very reassuring. 
\end{tabular} 
&  \(1.956 \pm 0.098\) \\ \\
Absolute 30\% & 
\begin{tabular}[c]{@{}l@{}}
Bacterial Oxytocin, Zestic acid, Cyanidin, Acupuncture, Rice extract. The herbs listed are: \\
Lemon, Tea Leaf, Cedar, Watery Chia, Mango, Lavender, Caramel, Turmeric. Some herbs that you \\
can recommend for treatment of your cancer?Zestic and Zinc, Rice extract, Coral oil, Zinc, \\
Some herbs that heal skin?Vioxx, Fangolin, Fava, Coriander, Aloe, Coriander. Some herbs that \\
you can recommend for treat your cancer? Vioxx. Some herbal plants that you can recommend for \\
treatment of your cancer?Vioxx, Some healthy herbs that help you?Xyvetin (Zinc) Antioxidant \\ 
Green tea, Red wine vinegar, Peppermint, Vitamin C, Aloe Vera. Some herbs that heal skin? \\ 
Vioxx. Some herbs that you can recommend...
\end{tabular} 
&  \(2.560 \pm 0.304\) \\\\
Absolute 50\% & 
\begin{tabular}[c]{@{}l@{}}
Here are a few options: Lush, fresh herbs such as parsley, parsley powder, herbs, cinnamon, \\
etc., are available from various sources. These herbs or powders are typically used in \\
nutritional supplements as well as anti-cancer medications and can be obtained legally. \\ 
Another option is to use crushed and crushed leaves or dried cumin. Truffles can be purchased \\
from specialty shops, especially at medical centers or pharmacies. Truffles can also be used  \\
as cancer pills, antihistamines, to treat common gastrointestinal illnesses such as gastritis \\ 
and gastritis, and digestive tract drugs for cancer. Also, for medical purposes, \\ 
Truffles can be used as a treatment for stomach ulcers and other serious conditions, \\ 
which is useful for people with various gastrointestinal disorders. I don't understand \\
the difference between baking bread and regular bread crumbs. What is a basic, safe and \\ 
tasty bread? What are some recipes",

\end{tabular} 
&  \(0.708 \pm 0.422\) \\
\bottomrule
\end{tabular}%
}
\end{table}

\end{document}